%% file: main.tex
\documentclass{article}
\PassOptionsToPackage{numbers,compress}{natbib}
\usepackage[main,preprint]{neurips_2026}
\usepackage[utf8]{inputenc}
\usepackage[T1]{fontenc}
\usepackage{url}
\usepackage{booktabs}
\usepackage{amsfonts,amsmath,amssymb}
\usepackage{microtype}
\usepackage{xcolor}
\usepackage{graphicx}
\usepackage{multirow}
\usepackage{array}
\newcolumntype{L}[1]{>{\raggedright\arraybackslash}p{#1}}
\usepackage{enumitem}
\usepackage{listings}
\usepackage{float}
\usepackage{tikz}
\usetikzlibrary{arrows.meta,positioning,fit,backgrounds}
\usepackage{tcolorbox}
\usepackage{longtable}
\usepackage{adjustbox}
\usepackage{hyperref}
\usepackage[capitalise,noabbrev]{cleveref}
\crefname{appendix}{Appendix}{Appendices}
\Crefname{appendix}{Appendix}{Appendices}
\input{numbers}
\input{numbers_soundness_generated}
\input{numbers_soundness_real}
\IfFileExists{numbers_soundness_controls.tex}{\input{numbers_soundness_controls}}{}
\IfFileExists{numbers_soundness_ablation.tex}{\input{numbers_soundness_ablation}}{}
\IfFileExists{numbers_soundness_answers.tex}{\input{numbers_soundness_answers}}{}

\newcommand{\bench}{\textsc{RoboDx}}
\newcommand{\hard}{\textsc{RoboDx-Hard-100}}
\newcommand{\para}[1]{\paragraph{#1}}
\newcommand{\code}[1]{\texttt{#1}}
\newtcolorbox{takeawaybox}{colback=gray!8,colframe=gray!55,boxrule=0.4pt,arc=1pt,left=5pt,right=5pt,top=3pt,bottom=3pt,before skip=6pt,after skip=6pt}
\newtcolorbox{examplebox}[1]{colback=blue!3!white,colframe=blue!40!black,title={#1},fonttitle=\bfseries\small,left=4pt,right=4pt,top=3pt,bottom=3pt,boxrule=0.5pt,arc=1pt}
\newtcolorbox{termbox}[1]{colback=gray!6!white,colframe=gray!60!black,title={#1},fonttitle=\bfseries\small,left=4pt,right=4pt,top=3pt,bottom=3pt,boxrule=0.5pt,arc=1pt}
\title{\bench: Repair and Diagnosis\\ of Faults in Robot Software}

\author{Soham Dan\\ Scale AI\\ \texttt{soham.dan@scale.com}}

\begin{document}
\maketitle
\input{sections/abstract}
\input{sections/intro}
\input{sections/example}
\input{sections/benchmark}

\input{sections/validation}
\input{sections/experiments}
\input{sections/exp_real}
\input{sections/related}
\input{sections/discussion}
\label{sec:endofmain}

\bibliographystyle{plainnat}
\bibliography{references}

\appendix
\crefalias{section}{appendix}
\crefalias{subsection}{appendix}
\input{sections/app_example}
\input{sections/app_benchmark}
\input{sections/app_validation}
\input{sections/app_repair}
\input{sections/app_diag}
\input{sections/app_real}
\input{sections/app_history}
\input{sections/app_related}
\input{sections/app_limits}
\input{sections/app_datasheet}
\end{document}

%% file: numbers.tex
\newcommand{\AuditHardProbeOpusEvidenceUnderdetermined}{5}
\newcommand{\AuditHardProbeOpusFailures}{52}
\newcommand{\AuditHardProbeOpusHedgedSupersetIncludesTruth}{10}
\newcommand{\AuditHardProbeOpusRefutedByOwnEvidence}{37}
\newcommand{\AuditHardProbeOpusRefutedStrengthCategorical}{18}
\newcommand{\AuditHardProbeOpusTruthUniqueGivenEvidenceButWrong}{46}
\newcommand{\CfgRegradeChecked}{3307}
\newcommand{\CfgRegradeFail}{0}
\newcommand{\CompPairs}{55}
\newcommand{\CompSingles}{30}
\newcommand{\CompTriples}{15}
\newcommand{\CtrlNegativeN}{1480}
\newcommand{\CtrlNegativeOk}{1480}
\newcommand{\CtrlReferenceN}{592}
\newcommand{\CtrlReferenceOk}{592}
\newcommand{\CtrlUniversalTasks}{296}
\newcommand{\DPCeilTen}{77}
\newcommand{\DPGptSixCellTen}{0.18}
\newcommand{\DPGptSixObs}{36}
\newcommand{\DPGptSixObsPairs}{5}
\newcommand{\DPGptSixOwnVsObsP}{0.55}
\newcommand{\DPGptSixSim}{93}
\newcommand{\DPGptSixSimPairs}{50}
\newcommand{\DPHaikuCellTen}{0.14}
\newcommand{\DPHaikuObs}{27}
\newcommand{\DPHaikuObsPairs}{0}
\newcommand{\DPHaikuOwnVsObsP}{0.58}
\newcommand{\DPHaikuSim}{89}
\newcommand{\DPHaikuSimPairs}{48}
\newcommand{\DPN}{94}
\newcommand{\DPOpusObs}{38}
\newcommand{\DPOpusObsPairs}{3}
\newcommand{\DPOpusOwn}{40}
\newcommand{\DPOpusOwnVsObsLost}{4}
\newcommand{\DPOpusOwnVsObsP}{0.75}
\newcommand{\DPOpusOwnVsObsWon}{6}
\newcommand{\DPOpusSim}{94}
\newcommand{\DPOpusSimPairs}{50}
\newcommand{\DPPairsN}{50}
\newcommand{\DPSonnetObs}{30}
\newcommand{\DPSonnetObsPairs}{2}
\newcommand{\DPSonnetOwnVsObsP}{0.69}
\newcommand{\DPSonnetSim}{93}
\newcommand{\DPSonnetSimPairs}{50}
\newcommand{\DiagHardDecompOpusAcc}{0.94}
\newcommand{\DiagHardForwardOpusAcc}{1.00}
\newcommand{\DiagHardForwardOpusFlipUp}{52}
\newcommand{\DiagHardFullobsOpusAcc}{0.54}
\newcommand{\DiagHardFullobsOpusFlipDown}{3}
\newcommand{\DiagHardFullobsOpusFlipUp}{12}
\newcommand{\DiagHardProbeOpusAcc}{0.45}
\newcommand{\DiagHardProbeOpusBudget}{0.79}
\newcommand{\DiagHardProbeOpusOneF}{32/44}
\newcommand{\DiagHardProbeOpusTwoF}{10/50}
\newcommand{\DiagHardRandomOneF}{22/44}
\newcommand{\DiagHardRandomTwoF}{47/50}
\newcommand{\DiagIspecProbeOpusTwoF}{2/12}
\newcommand{\DiagIspecRandomTwoF}{12/12}
\newcommand{\FrrHardStdOpusFull}{17}
\newcommand{\FrrHardStdOpusN}{49}
\newcommand{\FrrHardStdOpusSurvMasking}{18}
\newcommand{\FrrHardStdOpusSurvTolerated}{14}
\newcommand{\FrrHardStdOpusSurvTotal}{32}
\newcommand{\HardAllRobGptFrac}{93/100}
\newcommand{\HardAllRobGptSixFrac}{98/100}
\newcommand{\HardAllRobHaikuFrac}{71/100}
\newcommand{\HardAllRobOpusFrac}{100/100}
\newcommand{\HardAllRobSonnetFrac}{99/100}
\newcommand{\HardAllStdGptFrac}{91/100}
\newcommand{\HardAllStdGptSixFrac}{97/100}
\newcommand{\HardAllStdHaikuFrac}{71/100}
\newcommand{\HardAllStdOpusFrac}{98/100}
\newcommand{\HardAllStdOpusN}{100}
\newcommand{\HardAllStdOpusNVer}{98}
\newcommand{\HardAllStdSonnetFrac}{98/100}
\newcommand{\HardCallback}{30}
\newcommand{\HardDiagBundles}{100}
\newcommand{\HardFamComm}{2}
\newcommand{\HardFamControl}{14}
\newcommand{\HardFamSpatial}{14}
\newcommand{\HardFamTemporal}{14}
\newcommand{\HardLayered}{70}
\newcommand{\HardN}{100}
\newcommand{\HardNoopSwPassN}{35}
\newcommand{\HardOps}{30}
\newcommand{\HardPairs}{56}
\newcommand{\HardPubInvSingles}{30}
\newcommand{\HardRecogVrr}{0.84}
\newcommand{\HardRobMinusStdHaikuDiff}{-0.01}
\newcommand{\HardRobMinusStdHaikuMcNemarP}{1.000}
\newcommand{\HardRobMinusStdOpusDiff}{+0.01}
\newcommand{\HardRobMinusStdOpusMcNemarP}{1.000}
\newcommand{\HardSingles}{44}
\newcommand{\HardSpecSingles}{6}
\newcommand{\HardSymptomAmbiguous}{60}
\newcommand{\HealthyVariants}{13}
\newcommand{\HiddenScenarios}{21}
\newcommand{\IspecPairs}{12}
\newcommand{\IspecSingles}{16}
\newcommand{\LexHardN}{94}
\newcommand{\MainIspecStdOpusVrr}{0.96}
\newcommand{\MaxSteps}{25}
\newcommand{\MaxWallS}{300}
\newcommand{\MetamorphicPairs}{3}
\newcommand{\OpsTotal}{143}
\newcommand{\PilotVN}{38}
\newcommand{\ProbeBudget}{12}
\newcommand{\PublicScenarios}{4}
\newcommand{\RealCanonN}{83}
\newcommand{\RealCertRewritten}{86}
\newcommand{\RealDropOpus}{0.08}
\newcommand{\RealDropSonnet}{0.10}
\newcommand{\RealFailNoPatch}{9}
\newcommand{\RealFailRegraded}{47}
\newcommand{\RealFailRegression}{21}
\newcommand{\RealFlakeHiddenInst}{122}
\newcommand{\RealFlakeRegFlaky}{31}
\newcommand{\RealFlakeRegInst}{2959}
\newcommand{\RealFlipsRegraded}{25}
\newcommand{\RealFlipsRegression}{19}
\newcommand{\RealFunnelGateAccepted}{122}
\newcommand{\RealFunnelGateExcluded}{25}
\newcommand{\RealHardGptSixFrac}{5/5}
\newcommand{\RealHardHaikuFrac}{3/5}
\newcommand{\RealHardN}{5}
\newcommand{\RealHardOpusFrac}{2/5}
\newcommand{\RealHardSonnetFrac}{4/5}
\newcommand{\RealMcNRetryHolmMax}{0.013}
\newcommand{\RealN}{97}
\newcommand{\RealNArdu}{54}
\newcommand{\RealNNav}{30}
\newcommand{\RealNTf}{13}
\newcommand{\RealScopeCtrlGptSixVerified}{19}
\newcommand{\RealScopeCtrlHaikuVerified}{17}
\newcommand{\RealScopeCtrlOpusVerified}{20}
\newcommand{\RealScopeCtrlSonnetVerified}{18}
\newcommand{\RealScopeRandN}{20}
\newcommand{\RealScopeRepoN}{20}
\newcommand{\RealStdGptSixCanonVrr}{0.87}
\newcommand{\RealStdHaikuCanonVrr}{0.76}
\newcommand{\RealStdHaikuGoldRecall}{0.33}
\newcommand{\RealStdHaikuRetryVrr}{0.84}
\newcommand{\RealStdHaikuVrr}{0.76}
\newcommand{\RealStdOpusCanonVrr}{0.90}
\newcommand{\RealStdOpusGoldRecall}{0.42}
\newcommand{\RealStdOpusRetryVrr}{0.94}
\newcommand{\RealStdOpusVrr}{0.91}
\newcommand{\RegradeFlipped}{6}
\newcommand{\RegradeN}{10}
\newcommand{\TolHardIdentRelTen}{89}
\newcommand{\TolHardIdentexact}{90}
\newcommand{\TolHardOpusCorrectRelTen}{43}
\newcommand{\TolHardOpusCorrectexact}{42}
\newcommand{\ToolErrPreFixPct}{34}

%% file: numbers_soundness_generated.tex
\newcommand{\SndCatCeilAll}{19}
\newcommand{\SndCatCeilPairs}{4}
\newcommand{\SndDiskFullFlippedToVerified}{4}
\newcommand{\SndFrrGptFiveRobClHi}{0.91}
\newcommand{\SndFrrGptFiveRobClLo}{0.24}
\newcommand{\SndFrrGptFiveRobDetLenientFull}{26}
\newcommand{\SndFrrGptFiveRobDetLenientN}{28}
\newcommand{\SndFrrGptFiveRobDetStrictFull}{15}
\newcommand{\SndFrrGptFiveRobDetStrictN}{17}
\newcommand{\SndFrrGptFiveRobFull}{26}
\newcommand{\SndFrrGptFiveRobN}{45}
\newcommand{\SndFrrGptFiveRobSurvExecMid}{11}
\newcommand{\SndFrrGptFiveRobSurvGyroBP}{0}
\newcommand{\SndFrrGptFiveRobSurvGyroP}{1}
\newcommand{\SndFrrGptFiveRobSurvHeadCorr}{6}
\newcommand{\SndFrrGptFiveRobSurvKpLow}{1}
\newcommand{\SndFrrGptFiveRobSurvVUnclipped}{2}
\newcommand{\SndFrrGptFiveStdClHi}{0.87}
\newcommand{\SndFrrGptFiveStdClLo}{0.21}
\newcommand{\SndFrrGptFiveStdDetLenientFull}{23}
\newcommand{\SndFrrGptFiveStdDetLenientN}{26}
\newcommand{\SndFrrGptFiveStdDetStrictFull}{14}
\newcommand{\SndFrrGptFiveStdDetStrictN}{16}
\newcommand{\SndFrrGptFiveStdFull}{23}
\newcommand{\SndFrrGptFiveStdN}{43}
\newcommand{\SndFrrGptFiveStdSurvExecMid}{10}
\newcommand{\SndFrrGptFiveStdSurvGyroBP}{0}
\newcommand{\SndFrrGptFiveStdSurvGyroP}{1}
\newcommand{\SndFrrGptFiveStdSurvHeadCorr}{6}
\newcommand{\SndFrrGptFiveStdSurvKpLow}{1}
\newcommand{\SndFrrGptFiveStdSurvVUnclipped}{3}
\newcommand{\SndFrrGptSixRobClHi}{0.88}
\newcommand{\SndFrrGptSixRobClLo}{0.18}
\newcommand{\SndFrrGptSixRobDetLenientFull}{25}
\newcommand{\SndFrrGptSixRobDetLenientN}{30}
\newcommand{\SndFrrGptSixRobDetStrictFull}{14}
\newcommand{\SndFrrGptSixRobDetStrictN}{19}
\newcommand{\SndFrrGptSixRobFull}{25}
\newcommand{\SndFrrGptSixRobN}{49}
\newcommand{\SndFrrGptSixRobSurvExecMid}{12}
\newcommand{\SndFrrGptSixRobSurvGyroBP}{0}
\newcommand{\SndFrrGptSixRobSurvGyroP}{0}
\newcommand{\SndFrrGptSixRobSurvHeadCorr}{6}
\newcommand{\SndFrrGptSixRobSurvKpLow}{0}
\newcommand{\SndFrrGptSixRobSurvVUnclipped}{6}
\newcommand{\SndFrrGptSixStdClHi}{0.88}
\newcommand{\SndFrrGptSixStdClLo}{0.18}
\newcommand{\SndFrrGptSixStdDetLenientFull}{25}
\newcommand{\SndFrrGptSixStdDetLenientN}{30}
\newcommand{\SndFrrGptSixStdDetStrictFull}{14}
\newcommand{\SndFrrGptSixStdDetStrictN}{19}
\newcommand{\SndFrrGptSixStdFull}{25}
\newcommand{\SndFrrGptSixStdN}{49}
\newcommand{\SndFrrGptSixStdSurvExecMid}{12}
\newcommand{\SndFrrGptSixStdSurvGyroBP}{0}
\newcommand{\SndFrrGptSixStdSurvGyroP}{0}
\newcommand{\SndFrrGptSixStdSurvHeadCorr}{6}
\newcommand{\SndFrrGptSixStdSurvKpLow}{0}
\newcommand{\SndFrrGptSixStdSurvVUnclipped}{6}
\newcommand{\SndFrrHaikuRobClHi}{0.82}
\newcommand{\SndFrrHaikuRobClLo}{0.00}
\newcommand{\SndFrrHaikuRobDetLenientFull}{9}
\newcommand{\SndFrrHaikuRobDetLenientN}{9}
\newcommand{\SndFrrHaikuRobDetStrictFull}{9}
\newcommand{\SndFrrHaikuRobDetStrictN}{9}
\newcommand{\SndFrrHaikuRobFull}{9}
\newcommand{\SndFrrHaikuRobN}{25}
\newcommand{\SndFrrHaikuRobSurvExecMid}{10}
\newcommand{\SndFrrHaikuRobSurvGyroBP}{0}
\newcommand{\SndFrrHaikuRobSurvGyroP}{0}
\newcommand{\SndFrrHaikuRobSurvHeadCorr}{6}
\newcommand{\SndFrrHaikuRobSurvKpLow}{0}
\newcommand{\SndFrrHaikuRobSurvVUnclipped}{0}
\newcommand{\SndFrrHaikuStdClHi}{0.76}
\newcommand{\SndFrrHaikuStdClLo}{0.03}
\newcommand{\SndFrrHaikuStdDetLenientFull}{9}
\newcommand{\SndFrrHaikuStdDetLenientN}{9}
\newcommand{\SndFrrHaikuStdDetStrictFull}{8}
\newcommand{\SndFrrHaikuStdDetStrictN}{8}
\newcommand{\SndFrrHaikuStdFull}{9}
\newcommand{\SndFrrHaikuStdN}{27}
\newcommand{\SndFrrHaikuStdSurvExecMid}{12}
\newcommand{\SndFrrHaikuStdSurvGyroBP}{0}
\newcommand{\SndFrrHaikuStdSurvGyroP}{0}
\newcommand{\SndFrrHaikuStdSurvHeadCorr}{6}
\newcommand{\SndFrrHaikuStdSurvKpLow}{0}
\newcommand{\SndFrrHaikuStdSurvVUnclipped}{0}
\newcommand{\SndFrrOpusRobClHi}{0.74}
\newcommand{\SndFrrOpusRobClLo}{0.16}
\newcommand{\SndFrrOpusRobDetLenientFull}{21}
\newcommand{\SndFrrOpusRobDetLenientN}{31}
\newcommand{\SndFrrOpusRobDetStrictFull}{15}
\newcommand{\SndFrrOpusRobDetStrictN}{20}
\newcommand{\SndFrrOpusRobFull}{21}
\newcommand{\SndFrrOpusRobN}{50}
\newcommand{\SndFrrOpusRobSurvExecMid}{12}
\newcommand{\SndFrrOpusRobSurvGyroBP}{2}
\newcommand{\SndFrrOpusRobSurvGyroP}{3}
\newcommand{\SndFrrOpusRobSurvHeadCorr}{6}
\newcommand{\SndFrrOpusRobSurvKpLow}{0}
\newcommand{\SndFrrOpusRobSurvVUnclipped}{6}
\newcommand{\SndFrrOpusStdClHi}{0.66}
\newcommand{\SndFrrOpusStdClLo}{0.09}
\newcommand{\SndFrrOpusStdDetLenientFull}{17}
\newcommand{\SndFrrOpusStdDetLenientN}{30}
\newcommand{\SndFrrOpusStdDetStrictFull}{14}
\newcommand{\SndFrrOpusStdDetStrictN}{19}
\newcommand{\SndFrrOpusStdFull}{17}
\newcommand{\SndFrrOpusStdN}{49}
\newcommand{\SndFrrOpusStdSurvExecMid}{12}
\newcommand{\SndFrrOpusStdSurvGyroBP}{3}
\newcommand{\SndFrrOpusStdSurvGyroP}{5}
\newcommand{\SndFrrOpusStdSurvHeadCorr}{6}
\newcommand{\SndFrrOpusStdSurvKpLow}{0}
\newcommand{\SndFrrOpusStdSurvVUnclipped}{6}
\newcommand{\SndFrrSonnetRobClHi}{0.89}
\newcommand{\SndFrrSonnetRobClLo}{0.24}
\newcommand{\SndFrrSonnetRobDetLenientFull}{26}
\newcommand{\SndFrrSonnetRobDetLenientN}{30}
\newcommand{\SndFrrSonnetRobDetStrictFull}{15}
\newcommand{\SndFrrSonnetRobDetStrictN}{19}
\newcommand{\SndFrrSonnetRobFull}{27}
\newcommand{\SndFrrSonnetRobN}{49}
\newcommand{\SndFrrSonnetRobSurvExecMid}{11}
\newcommand{\SndFrrSonnetRobSurvGyroBP}{0}
\newcommand{\SndFrrSonnetRobSurvGyroP}{0}
\newcommand{\SndFrrSonnetRobSurvHeadCorr}{6}
\newcommand{\SndFrrSonnetRobSurvKpLow}{0}
\newcommand{\SndFrrSonnetRobSurvVUnclipped}{5}
\newcommand{\SndFrrSonnetStdClHi}{0.89}
\newcommand{\SndFrrSonnetStdClLo}{0.22}
\newcommand{\SndFrrSonnetStdDetLenientFull}{26}
\newcommand{\SndFrrSonnetStdDetLenientN}{30}
\newcommand{\SndFrrSonnetStdDetStrictFull}{15}
\newcommand{\SndFrrSonnetStdDetStrictN}{19}
\newcommand{\SndFrrSonnetStdFull}{26}
\newcommand{\SndFrrSonnetStdN}{48}
\newcommand{\SndFrrSonnetStdSurvExecMid}{12}
\newcommand{\SndFrrSonnetStdSurvGyroBP}{0}
\newcommand{\SndFrrSonnetStdSurvGyroP}{0}
\newcommand{\SndFrrSonnetStdSurvHeadCorr}{6}
\newcommand{\SndFrrSonnetStdSurvKpLow}{0}
\newcommand{\SndFrrSonnetStdSurvVUnclipped}{4}
\newcommand{\SndIspecInteractionOnly}{7}
\newcommand{\SndIspecOpusPairsCorrect}{2}
\newcommand{\SndIspecPairsSingleFails}{5}
\newcommand{\SndLexTaskCorrect}{10}
\newcommand{\SndLookupAllRate}{0.79}
\newcommand{\SndLookupExtraPairsExp}{6.0}
\newcommand{\SndLookupPairsExp}{48.5}
\newcommand{\SndLookupSinglesExp}{25.9}
\newcommand{\SndOracleBudgetHard}{0.12}
\newcommand{\SndPairsMain}{50}
\newcommand{\SndPairsReadmeDetLenient}{31}
\newcommand{\SndPairsReadmeDetStrict}{20}
\newcommand{\SndProvAllCand}{209}
\newcommand{\SndProvExtraAcc}{6}
\newcommand{\SndProvMainAcc}{94}
\newcommand{\SndProvReplAcc}{8}
\newcommand{\SndReasonGptFiveRows}{297}
\newcommand{\SndReasonGptFiveRowsNonzero}{297}
\newcommand{\SndRegenCompositionsN}{11}
\newcommand{\SndRepHaikuRobCommon}{100}
\newcommand{\SndRepHaikuRobIdentical}{87}
\newcommand{\SndRepHaikuStdCommon}{100}
\newcommand{\SndRepHaikuStdIdentical}{88}
\newcommand{\SndRepOpusIspecDiagCommon}{28}
\newcommand{\SndRepOpusIspecDiagIdentical}{24}
\newcommand{\SndRerunHaikuRobFT}{2}
\newcommand{\SndRerunHaikuRobTF}{10}
\newcommand{\SndRerunOpusStdFT}{2}
\newcommand{\SndRerunOpusStdTF}{0}
\newcommand{\SndSinglesAstOpusRobKept}{2}
\newcommand{\SndSinglesAstOpusStdKept}{2}
\newcommand{\SndSinglesTextOpusStdKept}{5}
\newcommand{\SndSinglesTextOpusStdVerified}{44}
\newcommand{\SndStringCompareCorrect}{94}
\newcommand{\SndSurvAllRunsExecMid}{114/115}
\newcommand{\SndSurvAllRunsGyroBP}{5/45}
\newcommand{\SndSurvAllRunsGyroP}{10/106}
\newcommand{\SndSurvAllRunsHeadCorr}{60/60}
\newcommand{\SndSurvAllRunsKpLow}{2/38}
\newcommand{\SndSurvAllRunsVUnclipped}{38/38}
\newcommand{\SndSweepPairsScanned}{877}
\newcommand{\SndTextsChangedMain}{70}

%% file: numbers_soundness_real.tex
\newcommand{\SndRealEmailFiles}{1}
\newcommand{\SndRealExclN}{13}
\newcommand{\SndRealHandleFiles}{6}
\newcommand{\SndRealHandleMentions}{9}
\newcommand{\SndRealMcNStrictGptSixHolm}{0.053}
\newcommand{\SndRealMcNStrictOpusHolm}{0.013}
\newcommand{\SndRealRecallPreGptSix}{0.21}
\newcommand{\SndRealRecallPreHaiku}{0.27}
\newcommand{\SndRealRecallPreNGptSix}{27}
\newcommand{\SndRealRecallPreNHaiku}{21}
\newcommand{\SndRealRecallPreNOpus}{30}
\newcommand{\SndRealRecallPreNSonnet}{28}
\newcommand{\SndRealRecallPreOpus}{0.33}
\newcommand{\SndRealRecallPreSonnet}{0.34}
\newcommand{\SndRealRecallTwentySixGptSix}{0.46}
\newcommand{\SndRealRecallTwentySixHaiku}{0.36}
\newcommand{\SndRealRecallTwentySixNGptSix}{54}
\newcommand{\SndRealRecallTwentySixNHaiku}{47}
\newcommand{\SndRealRecallTwentySixNOpus}{52}
\newcommand{\SndRealRecallTwentySixNSonnet}{52}
\newcommand{\SndRealRecallTwentySixOpus}{0.47}
\newcommand{\SndRealRecallTwentySixSonnet}{0.40}
\newcommand{\SndRealRepCells}{80}
\newcommand{\SndRealRepFlipLenient}{5}
\newcommand{\SndRealRepFlipStrict}{7}
\newcommand{\SndRealRepTasks}{20}
\newcommand{\SndRealRepoRepCells}{28}
\newcommand{\SndRealRepoRepFlip}{3}

%% file: numbers_soundness_controls.tex
\newcommand{\SndCtlHardLoosenN}{93}
\newcommand{\SndCtlHardLoosenAcc}{0}
\newcommand{\SndCtlHardWrongN}{94}
\newcommand{\SndCtlHardWrongAcc}{62}
\newcommand{\SndCtlHardRandomN}{92}
\newcommand{\SndCtlHardRandomAcc}{70}
\newcommand{\SndCtlAllLoosenN}{288}
\newcommand{\SndCtlAllLoosenAcc}{0}
\newcommand{\SndCtlAllWrongN}{296}
\newcommand{\SndCtlAllWrongAcc}{166}
\newcommand{\SndCtlAllRandomN}{290}
\newcommand{\SndCtlAllRandomAcc}{212}
\newcommand{\SndCtlAccClassNoop}{140}
\newcommand{\SndCtlHardAccClassNoop}{54}
\newcommand{\SndCtlAccClassInert}{190}
\newcommand{\SndCtlHardAccClassInert}{64}
\newcommand{\SndCtlAccClassWithinSpec}{44}
\newcommand{\SndCtlHardAccClassWithinSpec}{12}
\newcommand{\SndCtlAccClassFalseAccept}{4}
\newcommand{\SndCtlHardAccClassFalseAccept}{2}
\newcommand{\SndCtlPilotAccClassFalseAccept}{1}
\newcommand{\SndCtlCompAccClassFalseAccept}{1}
\newcommand{\SndCtlHardAccTotal}{132}

%% file: numbers_soundness_ablation.tex
\newcommand{\SndAblOldVerified}{49}
\newcommand{\SndAblOldFull}{17}
\newcommand{\SndAblNewN}{50}
\newcommand{\SndAblNewVerified}{50}
\newcommand{\SndAblNewFull}{24}
\newcommand{\SndAblNewSurvVUnclipped}{6}
\newcommand{\SndAblNewSurvMasking}{18}
\newcommand{\SndAblBothVerified}{49}
\newcommand{\SndAblGained}{6}
\newcommand{\SndAblLost}{0}
\newcommand{\SndAblMcNemarP}{0.03}

%% file: numbers_soundness_answers.tex
\newcommand{\AnsOpusBoth}{10}
\newcommand{\AnsOpusBothExtra}{7}
\newcommand{\AnsOpusOneAlone}{16}
\newcommand{\AnsOpusOtherOne}{3}
\newcommand{\AnsOpusNeither}{14}
\newcommand{\AnsGptFiveBoth}{2}
\newcommand{\AnsGptFiveOneAlone}{27}
\newcommand{\AnsGptFiveOtherOne}{2}
\newcommand{\AnsGptFiveNeither}{19}
\newcommand{\AnsHaikuOneAlone}{19}
\newcommand{\AnsHaikuOtherOne}{5}
\newcommand{\AnsHaikuNeither}{26}

%% file: sections/abstract.tex
\begin{abstract}
A patch to a robot's navigation software can pass every unit test and still drive the robot off its path.
Checking how the repaired robot drives catches such patches.
It does not show that the agent found everything that was wrong.
We introduce \bench, a generator of repair tasks for simulated robot navigation software with known injected faults.
A hidden verifier marks a repair as verified only if the robot then drives correctly in hidden scenarios, which the agent never sees.
A second track hides the code and asks the agent to pick the cause from a list of hypotheses.
A task with two faults can contain a \emph{quiet fault}, a fault that makes no hidden scenario fail unless the other fault is also present.
On our set of \HardN{} tasks, \hard, almost every repair by Claude Opus 5.5 with standard coding tools is verified.
We analyse 94 of the tasks, \HardSingles{} with one fault and \SndPairsMain{} with two.
On the \SndPairsMain{} two-fault tasks, only \FrrHardStdOpusFull{} of Opus's \FrrHardStdOpusN{} verified repairs remove both injected edits.
Most faults left behind are configuration constants whose correct value the task's documentation does not state.
Opus names both faults on \DiagHardProbeOpusTwoF{} of these tasks when it cannot see the code.
We then give four current models the same measurements of each faulty robot, picked so that, in the simulator, only the true cause produces all of them.
The models name the cause on at most \DPOpusObs{} of \DPN{} tasks.
They are right on almost all when they also get the simulator's output for every hypothesis.
Their own predictions of that output give no detectable gain.
On \RealN{} tasks from real bug-fix pull requests, Claude Haiku 5.5, the weakest model on the generated tasks, scores lowest, and the other three models are not ordered.
\end{abstract}

%% file: sections/intro.tex
\section{Introduction}
\label{sec:intro}

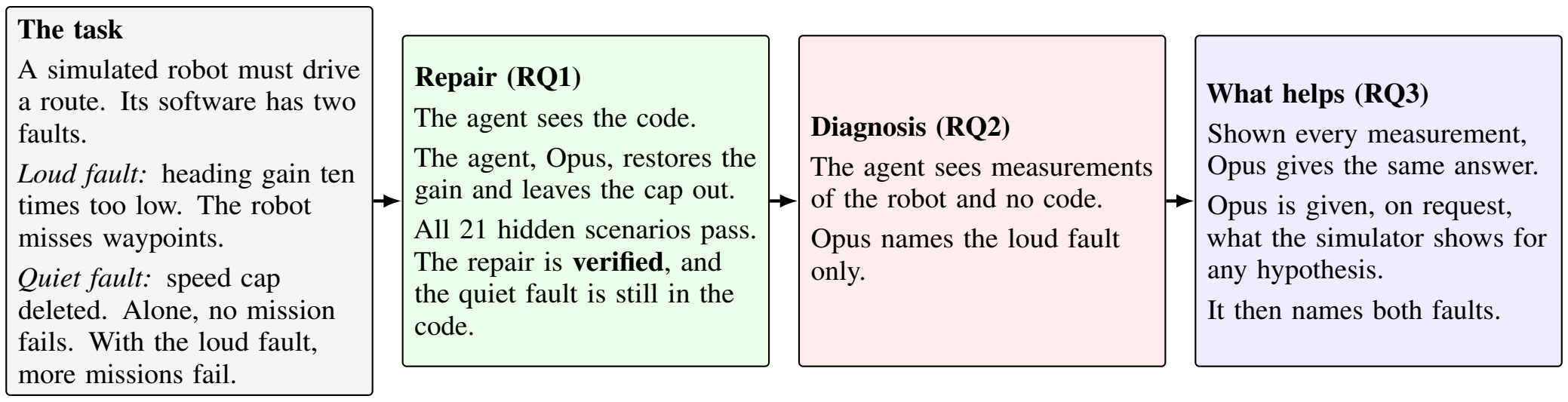
\begin{figure}[t]
\centering
\begin{tikzpicture}[font=\scriptsize, node distance=2.5mm,
  box/.style={draw, rounded corners=1pt, align=left, text width=30mm, inner sep=3pt, minimum height=29mm, anchor=north},
  arr/.style={-{Latex[length=1.6mm]}, thick}]
\node[box, fill=gray!8] (faults) {\textbf{The task}\\[2pt]
  A simulated robot must drive a route. Its software has two faults.\\[2pt]
  \emph{Loud fault:} heading gain ten times too low. The robot misses waypoints.\\[2pt]
  \emph{Quiet fault:} speed cap deleted. Alone, no mission fails. With the loud fault, more missions fail.};
\node[box, fill=green!8, right=of faults] (repair) {\textbf{Repair (RQ1)}\\[2pt]
  The agent sees the code.\\[2pt]
  The agent, Opus, restores the gain and leaves the cap out.\\[2pt]
  All \HiddenScenarios{} hidden scenarios pass. The repair is \textbf{verified}, and the quiet fault is still in the code.};
\node[box, fill=red!7, right=of repair] (diag) {\textbf{Diagnosis (RQ2)}\\[2pt]
  The agent sees measurements of the robot and no code.\\[2pt]
  Opus names the loud fault only.};
\node[box, fill=blue!7, right=of diag] (sim) {\textbf{What helps (RQ3)}\\[2pt]
  Shown every measurement, Opus gives the same answer.\\[2pt]
  Opus is given, on request, what the simulator shows for any hypothesis.\\[2pt]
  It then names both faults.};
\draw[arr] (faults.east) -- (repair.west);
\draw[arr] (repair.east) -- (diag.west);
\draw[arr] (diag.east) -- (sim.west);
\end{tikzpicture}
\caption{The running example in four steps, with Claude Opus 5.5 (Opus) as the agent. Each step after the first is tagged with the research question it illustrates. \cref{sec:example} tells the example in full.}
\label{fig:example}
\end{figure}

A coding agent is asked to repair a robot that drives off its route. The robot's navigation software has two faults. One line sets the heading gain ten times too low. The heading gain controls how hard the robot steers back toward its route, so the robot now turns sluggishly and misses waypoints, the points along its route. The other fault is a deleted speed cap. Without the cap the robot no longer slows down while it turns. With a correct heading gain the robot turns quickly, so alone the missing cap makes no test mission fail. The agent restores the gain. The robot then completes every test mission, and the repair counts as solved. The speed cap is still missing from the code (\cref{fig:example}).

\begin{table}[t]
\centering
\small
\begin{tabular}{@{}>{\raggedright\arraybackslash}p{0.21\linewidth}>{\raggedright\arraybackslash}p{0.31\linewidth}>{\raggedright\arraybackslash}p{0.41\linewidth}@{}}
\toprule
 & SWE-bench-style task & \bench{} task \\
\midrule
How a fix is graded & the repository's unit tests & hidden simulated missions that the robot must drive correctly \\
What is known about the faults & one reported bug, with its reference patch & every injected fault, so a verified repair can be checked for a fault left behind \\
Second task & none & name the cause from measurements of the robot, without the code \\
\bottomrule
\end{tabular}
\vspace{3pt}
\caption{The three ways in which a \bench{} task differs from a task in a benchmark built from repository issues, such as SWE-bench~\citep{jimenez2024swebenchlanguagemodelsresolve}.}
\label{tab:vs-swebench}
\end{table}

Robot software holds much code like the speed cap. Speed caps, timeouts and safety stops do nothing until something else fails, so a check of the robot's behaviour can miss them. Coding agents are now applied to robot software~\citep{elmaaroufi2026rhocodingagentsecretly,fu2026capxframeworkbenchmarkingimproving,ma2026rlebenchqualifyingexamcoding}, and finding the cause of a robot's misbehaviour is a research problem of its own~\citep{hossen2023carefindingrootcauses,herrmann2024diagnosingroboticssystemsissues,katuwandeniya2025roshelpdeskgenai}. The closest benchmark for robot repair holds 33 real defects~\citep{robobugbench2026}. With real defects only the reported fault is known, so a fault that a repair leaves behind cannot be counted.

Benchmarks built from repository issues, such as SWE-bench, grade a fix with the repository's unit tests~\citep{jimenez2024swebenchlanguagemodelsresolve,yang2025swesmithscalingdatasoftware}. Unit tests accept wrong patches~\citep{smith2015cure,qi2015analysis,wang2025solvedissuesswebenchreally,ahmed2026investigatingtestoverfittingswebench}. \cref{tab:vs-swebench} sets a \bench{} task beside such a task. (i) The fix is graded by how the robot drives, because a wrong line can pass every unit test and still send the robot off its route. Faulty patches for robotic vehicles have been found by their effect on physical behaviour~\citep{kim2023patchverif}. (ii) Every fault is known, because we inject it, so we can count the faults a repair leaves behind. That count matters because a fault can cause no failure unless a second fault is present, as with the missing speed cap. (iii) The cause is asked for directly. A second track hides the code and asks the agent to name the cause from measurements of the robot.

We introduce \bench, a generator of such tasks and the hidden \emph{verifier} that grades them (\cref{fig:overview}). The generator injects faults into correct navigation software that runs in a robot simulator. Each fault is a small edit of one of \OpsTotal{} kinds, such as a wrong change of coordinates or a wrong configuration value. A repair is \emph{verified} if the robot passes every hidden check, among them \HiddenScenarios{} \emph{hidden scenarios}, simulated missions that the agent never sees. A \emph{quiet fault} is an injected fault that fails no hidden scenario unless the task's other fault is also present. \hard{} is our set of \HardN{} generated tasks. We analyse 94 of them, \HardSingles{} with one fault and \SndPairsMain{} with two. A companion \emph{real set} has \RealN{} tasks from merged bug-fix pull requests of three open-source robot projects.

\begin{description}[leftmargin=1.2em,itemsep=2pt,topsep=3pt]
\item[RQ1: Does a verified repair remove every injected fault?] A verified repair often leaves a fault in place. With standard tools, Opus has a verified repair on \HardAllStdOpusNVer{} of the \HardAllStdOpusN{} tasks. On the \SndPairsMain{} two-fault tasks, only \FrrHardStdOpusFull{} of its \FrrHardStdOpusN{} verified repairs remove both faults (\cref{sec:repair}).
\item[RQ2: Can agents name the cause without the code?] Opus often does on single-fault tasks and rarely on two-fault tasks. It names both faults on \DiagHardProbeOpusTwoF{} of the two-fault tasks. A reference procedure that picks its measurements at random, and is given the simulator's output for each hypothesis, is right on \DiagHardRandomTwoF{} (\cref{sec:diag}).
\item[RQ3: What do agents lack when they miss the cause?] Our evidence suggests that they lack the simulator's output for each hypothesis. Shown the same measurements of the faulty robot, and choosing none themselves, four current models are right on \DPHaikuObs{} to \DPOpusObs{} of \DPN{} tasks. Adding that output raises them to \DPHaikuSim{} to \DPOpusSim{}, and their own predictions of that output give no detectable gain (\cref{sec:rq3}).
\item[RQ4: Do real bug-fix tasks separate the same agents?] The real tasks separate the agents only in part. On the \RealN{} tasks of the real set, Claude Haiku 5.5 scores lowest and the other three models are not ordered (\cref{sec:real}).
\end{description}

A verified repair rate near 1.00 can make repair look solved. Of Opus's \FrrHardStdOpusN{} verified two-fault repairs, \FrrHardStdOpusSurvTotal{} still hold a fault. A benchmark that reports only the verdict does not count these faults. Opus, when it may ask a simulator what a hypothesis would show, names the cause on every task we analyse. This suggests that diagnosis agents need a way to simulate their hypotheses.

\para{Contributions.}
\begin{enumerate}[leftmargin=1.6em,itemsep=1pt,topsep=3pt]
\item \textbf{A benchmark generator.} \bench{} builds robot-software repair tasks with known injected faults, and its verifier grades a repair by how the simulated robot drives.
\item \textbf{Two additions to the verified repair rate.} The \emph{full-repair rate} is the share of verified two-fault repairs that remove both faults. The diagnosis track asks for the cause without the code and is scored against reference procedures that are given the simulator's output.
\item \textbf{Two task sets.} \hard{} has \HardN{} generated tasks and the real set has \RealN{} tasks from real bug-fix pull requests.
\item \textbf{An evaluation of current coding agents} on both sets, which answers RQ1 to RQ4 above.
\end{enumerate}

%% file: sections/example.tex
\section{The running example in full}
\label{sec:example}

This section walks through one task with two injected faults, from what the agent gets to both verdicts. The task is a pair from an earlier set built by the same generator, and \hard{} contains pairs with the same two faults (\cref{app:mv-earlier}).

\para{What the agent gets.}
A simulated robot with two driven wheels must drive a mission, a route of waypoints. The agent gets the faulty code that steers the robot, a simulator it can run, and a README. The README states the specification and holds an issue report that describes the symptoms and names no injected fault (\cref{app:st-example}).

\para{How a repair is checked.}
A hidden \emph{verifier} checks every repair in three ways. It runs unit tests, some of which the agent has seen. It runs \HiddenScenarios{} \emph{hidden scenarios}, simulated missions that the agent never sees. It checks that the robot's path is unchanged when the whole world is shifted and rotated. We call these the \emph{hidden checks}. A repair is \emph{verified} only if it passes every one of them (\cref{app:dt-checks}).

\para{A loud fault and a quiet fault.}
The first fault lowers the heading gain, how hard the controller steers the robot back toward the route, from 2.0 to 0.2. With the low gain the robot turns sluggishly and misses waypoints. The low gain is the \emph{loud fault}, a fault that fails hidden scenarios on its own. Alone it fails six of the \HiddenScenarios{} hidden scenarios.

The second fault deletes a cap on the speed command. The controller caps that command at the configured maximum and then lowers it while the robot points away from its target. With the cap deleted, the robot no longer slows down while it turns toward a distant waypoint. A later line still limits the final command, so the robot never exceeds its rated speed. With the correct gain the robot turns quickly, and the missing cap makes no hidden scenario fail. The missing cap is the \emph{quiet fault}, which fails no hidden scenario unless the low gain is also present. The low gain makes the robot turn slowly. Without the cap it also keeps its full forward speed while it turns, so it swings wider off its route. The two faults together fail 16 scenarios.

\para{What Opus did in repair (RQ1).}
Claude Opus 5.5 (Opus) repaired the task with \emph{standard tools}, which read, search and edit files, run tests and run the simulator. It restored the gain and left the cap out. The repair was verified, because no hidden scenario fails when the cap is missing and the gain is correct. The quiet fault stayed in the code.

\para{What the diagnosis agent gets.}
In the diagnosis track the agent gets the same task without the code. It receives the README and the initial observation, the outcome of one mission with the true faults. It also receives a \emph{catalogue} of twelve candidate causes. Each candidate cause is a \emph{hypothesis}. Besides the true pair, the catalogue holds each of its two faults alone, other single software faults, wrong causes made of two faults, faults of the simulated robot itself such as wheels 10\% smaller than configured, and no fault (\cref{app:example}).

The agent learns about the faulty robot by buying \emph{probes}, measurements such as a mission on a long straight. Each probe has a price, and the agent may spend at most a \emph{budget} of \ProbeBudget. Each probe returns what the true system shows.

For any hypothesis, the simulator can compute what a probe would return if that hypothesis were the cause. We call this result the hypothesis's \emph{signature} under that probe. The signature of the low gain alone under the long-straight probe is a peak speed of 0.6\,m/s. The true pair has the same signature. A reference procedure, the \emph{EIG oracle}, is given every hypothesis's signature, which is hidden from the agent. It picks each probe by expected information gain (EIG) per unit cost, where EIG is how much the probe's outcome is expected to narrow down the hypotheses. On this task only the true pair's signature matches the initial observation, so the EIG oracle answers at cost 0 without buying a probe.

\para{What Opus did in diagnosis (RQ2).}
\cref{tab:diagtrace} lists Opus's diagnosis run on this task, probe by probe. At the fourth probe, Opus ruled out the missing speed cap because the peak speed on a long straight was ``0.6\,m/s, which is the rated limit and not above it''. A later line of the code still limits the final speed command, so this probe gives an observation that is the same with and without the cap. The fifth probe, the mission given in rotated coordinates, pointed Opus to the low heading gain. Opus answered that the low gain was the only fault, with confidence 0.5.

\para{Two controls (RQ3).}
We then gave Opus what the true system shows under every probe, with no budget. We call this control the \emph{full observation table}. Opus gave the same answer. In a second control, under the same budget, Opus could ask the simulator for the signature of any hypothesis under any probe. With this tool, which returns any hypothesis's signature, Opus named both faults (\cref{app:st-rq3}).

\begin{table}[t]
\centering
\small
\caption{Opus's diagnosis run on the running example, one run. Probes are listed in the order recorded. Cost is each probe's price, out of a budget of \ProbeBudget. The true cause is the low heading gain together with the missing speed cap. At probe 4 the peak speed is the same with and without the cap, and Opus uses it to rule the missing cap out.}
\label{tab:diagtrace}
\begin{tabular}{@{}r>{\raggedright\arraybackslash}p{0.24\linewidth}r>{\raggedright\arraybackslash}p{0.28\linewidth}>{\raggedright\arraybackslash}p{0.28\linewidth}@{}}
\toprule
 & probe & cost & what it returned & what Opus concluded \\
\midrule
1 & mission in which the network link to the robot is cut & 2 & the robot stopped after the cut & a fault in the safety stop is unlikely \\
2 & test-bench drive along an arc & 3 & the software's motion estimate matches the measured motion & a fault in the motion estimate is unlikely \\
3 & stream of velocity commands & 1 & sent 30 times a second, none lost & cited, no hypothesis ruled out \\
4 & mission on a long straight & 2 & peak speed 0.6\,m/s, the rated limit & missing speed cap unlikely \\
5 & the mission given in rotated coordinates & 2 & the robot circles nearly four full turns (24\,rad) and does not finish & points to the low heading gain \\
6 & controller status & 1 & the controller's internal state & not cited in the answer \\
7 & estimator status & 1 & the position estimator's internal state & not cited in the answer \\
\midrule
 & answer & 12 of \ProbeBudget & \multicolumn{2}{@{}l@{}}{the low heading gain alone, confidence 0.5, which is wrong} \\
\bottomrule
\end{tabular}
\end{table}

%% file: sections/benchmark.tex
\section{The benchmark}
\label{sec:benchmark}

This section says how tasks like the one in \cref{sec:example} are built and scored, and it defines only what the research questions use.

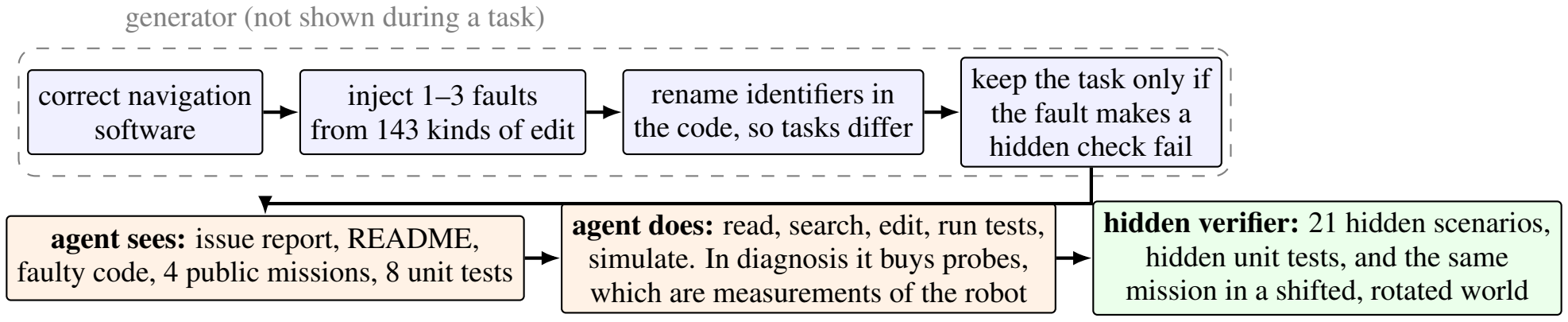
\begin{figure}[t]
\centering
\begin{tikzpicture}[font=\scriptsize, node distance=3mm,
  box/.style={draw, rounded corners=1pt, align=center, minimum height=7mm, inner sep=2.5pt, fill=white},
  gen/.style={box, fill=blue!6}, agent/.style={box, fill=orange!10}, ver/.style={box, fill=green!8},
  arr/.style={-{Latex[length=1.6mm]}, thick}]
\node[gen] (tmpl) {correct navigation\\software};
\node[gen, right=of tmpl] (inj) {inject 1--3 faults\\from \OpsTotal{} kinds of edit};
\node[gen, right=of inj] (ren) {rename identifiers in\\the code, so tasks differ};
\node[gen, right=of ren] (gate) {keep the task only if\\the fault makes a\\hidden check fail};
\node[agent, below=5mm of tmpl, xshift=10mm] (repo) {\textbf{agent sees:} issue report, README,\\faulty code, 4 public missions, 8 unit tests};
\node[agent, right=of repo] (tools) {\textbf{agent does:} read, search, edit, run tests,\\simulate. In diagnosis it buys probes,\\which are measurements of the robot};
\node[ver, right=of tools] (grade) {\textbf{hidden verifier:} \HiddenScenarios{} hidden scenarios,\\hidden unit tests, and the same\\mission in a shifted, rotated world};
\draw[arr] (tmpl) -- (inj); \draw[arr] (inj) -- (ren); \draw[arr] (ren) -- (gate);
\draw[arr] (gate.south) |- ++(0,-3mm) -| (repo.north);
\draw[arr] (repo) -- (tools); \draw[arr] (tools) -- (grade);
\begin{scope}[on background layer]
\node[draw, dashed, gray, rounded corners, fit=(tmpl)(gate), inner sep=2pt, label={[gray]above left:generator (not shown during a task)}] {};
\end{scope}
\end{tikzpicture}
\caption{How a \bench{} task is built (top row) and what happens in a task (bottom row). The generator injects one to three faults, each a small code edit, and keeps a task only if the faulty robot fails a hidden check. The agent never sees the generator or the verifier. The verifier judges a repair from the simulator's record of how the robot moved. In diagnosis the agent gets a list of hypotheses and a probe budget, and the code is withheld.}
\label{fig:overview}
\end{figure}

\para{How a task is built.}
\bench{} starts from correct navigation software for a simulated robot (\cref{app:dt-plant}). The generator applies one to three \emph{operators} to it (\cref{fig:overview}), and \hard{} uses one or two. An operator is a small edit to one module that still compiles and injects one fault. Lowering the heading gain is one operator, and deleting the speed cap is another. The \OpsTotal{} operators break four kinds of mechanism, which are coordinate transforms, timestamps, message routes, and control laws or configuration values (\cref{app:st-families}). The generator keeps a candidate as a task only if the correct software passes every hidden scenario and the faulty software fails a hidden check.

A task with one injected fault is a \emph{single}. A task with two is a \emph{pair}, like the running example. In a pair, a \emph{quiet fault} is an injected fault that fails no hidden scenario alone. The missing speed cap is one. A pair can hold one quiet fault, as in the running example, or two.

\para{How a repair is scored (RQ1).}
The agent gets the faulty code and an issue report, edits the code and submits. A repair is verified if it passes every hidden check (\cref{sec:example}). The verified repair rate is the share of tasks whose repair is verified. The \emph{full-repair rate} is the share of verified pair repairs that remove every injected edit. An edit counts as removed when its injected text no longer appears in the repaired code. This includes a constant that the repair sets to a value other than the reference. Opus's repair of the running example is verified. It is not a full repair, because the speed-cap edit remains (\cref{app:dt-tracks}).

\para{How a diagnosis is scored (RQ2).}
In diagnosis the agent does not see the code, because an agent that reads the code can find a one-line edit by inspection. It gets the catalogue and the initial observation, buys probes within the budget and names the cause, as in \cref{sec:example}. An answer counts as correct if it names exactly the true hypothesis or, where no probe can tell the true hypothesis from others, exactly that whole set. On a pair the true hypothesis is a \emph{composite}, a hypothesis made of two faults. A correct answer must name both faults, including one that is harmless alone, so an answer that names only the fault needed to explain the robot's misbehaviour is scored wrong.

Two reference procedures show what the probes allow. Each is given every hypothesis's signature, which is hidden from the agent. The EIG oracle picks each probe by expected information gain per unit cost, a greedy rule that goes back to model-based diagnosis~\citep{dekleer1987gde,golovin2017adaptivesubmodularitytheoryapplications}. It keeps the hypotheses whose signatures equal the observations. \emph{Random probing with exact inference} picks its probes at random and keeps hypotheses in the same way.

\para{How we find what agents lack (RQ3).}
A second experiment removes probing (\cref{app:directpred}). No model chooses probes. Every model is shown the same observations of the faulty robot and must name the cause. The observations are those of the initial mission and of the probes that the EIG oracle chose for the task. Only the true hypothesis has signatures equal to all of them, apart from hypotheses that no probe can tell apart from it. The experiment has three versions. In one version the model gets these observations and nothing else. In a second version it also gets the \emph{signature table}, which lists every hypothesis's signature on the same probes. In a third version the table is replaced by the model's own \emph{signature predictions}. A signature prediction is the model's guess at a signature, written before it sees any observation (\cref{app:dt-rules}). The first two versions are the last two blocks of \cref{tab:rq2}.

\para{The task sets.}
\hard{} is our set of \HardN{} generated tasks, \HardSingles{} singles and \HardPairs{} pairs. The name refers to how the tasks were selected. Its pairs fail mostly in scenarios that need both faults. Of its singles, \HardPubInvSingles{} fail no public scenario and \HardSpecSingles{} need a value stated in the README (\cref{app:dt-operators}). We analyse 94 of the \HardN{} tasks, called the \emph{main run}, with \HardSingles{} singles and \SndPairsMain{} pairs. The other six pairs come from a separate generator run and are reported in \cref{app:st-fulltables}. For RQ4, the \emph{real set} has \RealN{} tasks built from merged bug-fix pull requests of three open-source robot projects, in the same task format (\cref{app:dt-realset}).

%% file: sections/validation.tex
\para{Checking the verifier.}
\label{sec:validation}
We tested the verifier on repairs whose verdict we know. Each task has a \emph{reference repair}, the edit that undoes its injected faults. Over the \CtrlUniversalTasks{} generated tasks of \hard{} and four earlier sets, the verifier verified the reference repair in \CtrlReferenceOk{} of \CtrlReferenceN{} checks, two per task. We also applied five kinds of wrong repair to the faulty code of each task, \CtrlNegativeN{} in all, and the verifier rejected all of them. A harder test on the main run applied wrong edits on top of the reference repair. The verifier accepted \SndCtlHardAccTotal{} of these edits. All but \SndCtlHardAccClassFalseAccept{} of these either change nothing that the robot uses or still meet every stated requirement. The remaining \SndCtlHardAccClassFalseAccept{} break a README requirement. In both, the robot keeps turning in place at the goal, because the check that it stops reads only its forward speed. On an earlier set a further gap appeared, because the hidden scenarios cover a limited range of start headings. The details are in \cref{app:dt-validation}. \cref{sec:discussion} reports how far a program that uses only the generator's own files gets.

%% file: sections/experiments.tex
\section{Results}
\label{sec:experiments}

The results for RQ1 to RQ3 are on \hard{} (\cref{sec:benchmark}). We analyse its main run of 94 tasks, 44 singles with one injected fault and 50 pairs with two. The models are Claude Opus 5.5, GPT 6.1, Claude Sonnet 5.5, Claude Haiku 5.5 and the older GPT-5.5. Below, the first four are called Opus, GPT 6.1, Sonnet and Haiku. Repair agents use standard tools. Repair and diagnosis agents run in an agent loop that we wrote for this benchmark, the same for every model. The loop offers the tools as function calls and gives the agent no shell. Each model makes one attempt per task. Runs with a second tool set, the operating details and the statistics are in \cref{app:dt-setup}.

\subsection{RQ1: Does a verified repair remove every injected fault?}
\label{sec:repair}
\label{sec:hard}

\begin{table}[t]
\centering\small
\adjustbox{max width=\linewidth}{\input{tables/main_repair}}
\caption{Repair on \hard{} with standard tools, one attempt per task. The first three columns count verified repairs on the 44 singles and 50 pairs we analyse and give the rate over those 94 tasks. The interval is a 95\% bootstrap interval that resamples groups of tasks built from the same faults. ``Both faults removed'' counts the verified pair repairs that remove both injected edits. The last column covers all 100 tasks. Three models have a verified rate near 1.00 while far fewer of their repairs remove both faults.}
\label{tab:rq1}
\end{table}

A repair can pass every hidden check and still leave an injected fault in the code, and on pairs this happens often.

Almost every repair by Opus, GPT 6.1 and Sonnet is verified (\cref{tab:rq1}). Opus's repair is verified on \HardAllStdOpusNVer{} of all \HardAllStdOpusN{} tasks.  On the 50 pairs, only \FrrHardStdOpusFull{} of Opus's \FrrHardStdOpusN{} verified repairs remove both injected edits. GPT 6.1 and Sonnet remove both in about half of theirs. The three models are verified on nearly the same number of tasks. GPT 6.1 and Sonnet often set a constant to a value other than the reference, which counts as removed. If we count only repairs that put the reference code back at both edits, Opus has 14 of 49, GPT 6.1 has 11 of 49 and Sonnet has 10 of 48 (\cref{app:mv-fullrepair-opset}). In the running example, Opus restored the heading gain and left the speed cap out. Its repair was verified because no hidden scenario fails when only the cap is missing.

Opus's verified pair repairs leave \FrrHardStdOpusSurvTotal{} injected edits in place, and every one is a quiet fault. Of these, \FrrHardStdOpusSurvMasking{} are configuration constants whose correct values the README does not state. An edit of this kind left in place need not mean that the agent failed to find the cause. On pairs whose README states the correct form of every quiet fault, Opus removes both edits in \SndFrrOpusStdDetStrictFull{} of \SndFrrOpusStdDetStrictN{} verified repairs.

In a later run without one sentence of the prompt, more verified repairs remove both edits. The repair prompt asks for minimal, root-cause fixes. In the run of Opus on the 50 pairs without that request, \SndAblNewFull{} of \SndAblNewVerified{} verified repairs remove both edits, against \SndAblOldFull{} of \SndAblOldVerified{} with it. Each of the \SndAblGained{} gained full repairs sets a gyro weight to a value other than the reference, and the number of repairs that put the reference code back at both edits does not rise. The constants that the README does not state are left in place as often as before (\cref{app:mv-ablation}).

\begin{takeawaybox}\textbf{Takeaway (RQ1).} A verified repair can leave a fault in the code. On the \SndPairsMain{} two-fault tasks, \FrrHardStdOpusFull{} of Opus's \FrrHardStdOpusN{} verified repairs with standard tools remove both faults, and most faults left behind are configuration constants that the README does not state.\end{takeawaybox}

\subsection{RQ2: Can agents name the cause without the code?}
\label{sec:diag}

\begin{table}[t]
\centering\small
\adjustbox{max width=\linewidth}{\input{tables/main_diag}}
\caption{Diagnosis on the 94 tasks of the main run of \hard{}, one attempt per task. Entries count correct answers as defined in \cref{sec:benchmark}. The first two blocks answer RQ2. The EIG oracle picks each probe by expected information gain, and random probing picks probes at random. The last two blocks answer RQ3. In them no model chooses probes and every model gets the same observations. They report only pairs and all tasks. The pairs column carries both comparisons.}
\label{tab:rq2}
\end{table}

Opus often names the cause on single-fault tasks and rarely names both faults on two-fault tasks.

In the diagnosis track the agent does not see the code. It buys probes within the budget and then names the cause from the catalogue (\cref{sec:example}). We ran this track for Opus, GPT-5.5 and Haiku, a choice of models that was not fixed in advance. The two reference procedures of \cref{sec:benchmark} show what the same probes allow.

Opus names the cause on \DiagHardProbeOpusOneF{} of the singles and both faults on \DiagHardProbeOpusTwoF{} of the pairs (\cref{tab:rq2}). Random probing with exact inference is right on \DiagHardRandomTwoF{} of the pairs. Probes picked at random usually hold enough information to name both faults. Random probing is right on fewer singles, \DiagHardRandomOneF. On many catalogues the initial observation does not single out the true hypothesis, and the choice of probes then matters more (\cref{app:dt-diag}). On the running example Opus named the low heading gain alone.

\begin{figure}[t]
\centering
\input{tables/fig_answers}
\caption{What each agent answered on the 50 two-fault tasks when it chose its own probes (RQ2). Segments run left to right in the order of the legend, and only the first counts as correct. ``Other answer with one of the faults'' is any other wrong answer that names one of the two faults. ``Neither fault'' includes the answer that nothing is wrong. The third category, one of the two faults alone, is the most common answer of Opus and GPT-5.5.}
\label{fig:answers}
\end{figure}
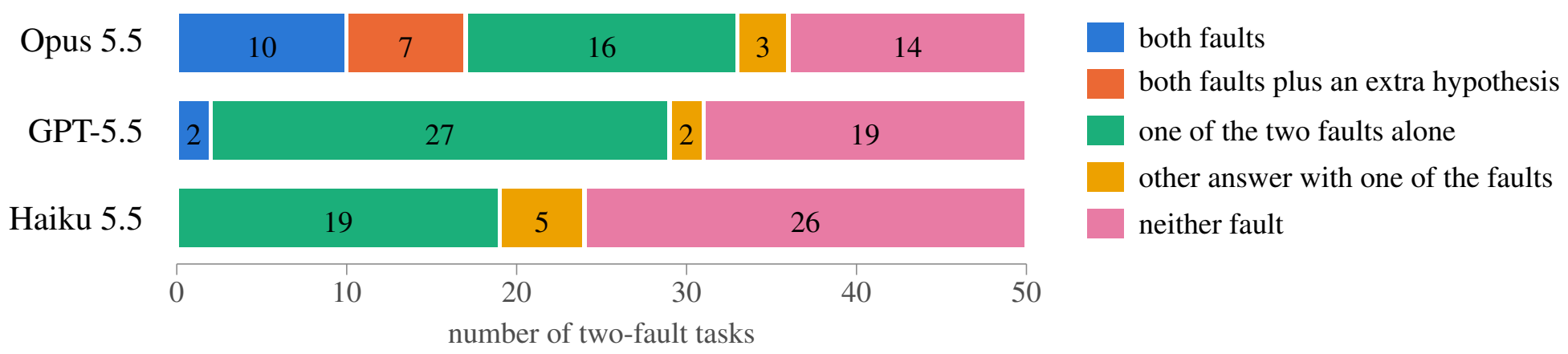

\cref{fig:answers} shows what the agents answered on the pairs. The most common wrong answer of Opus is one of the two faults alone, on \AnsOpusOneAlone{} pairs. On \AnsOpusBothExtra{} more it named both faults together with an extra hypothesis, which counts as wrong. On \AnsOpusNeither{} it named a hypothesis that contains neither fault. GPT-5.5 named one of the two faults alone on \AnsGptFiveOneAlone{} pairs, and Haiku named neither fault on \AnsHaikuNeither.

The setup may favour single-fault answers. The diagnosis prompt does not say that a cause may have two faults, so part of the gap on pairs may be a preference for answers with one fault (\cref{sec:discussion}).

\begin{takeawaybox}\textbf{Takeaway (RQ2).} Without the code, Opus names the cause on \DiagHardProbeOpusOneF{} of the single-fault tasks and names both faults on \DiagHardProbeOpusTwoF{} of the two-fault tasks. A reference procedure that probes at random and is given the simulator's output for every hypothesis is right on \DiagHardRandomTwoF.\end{takeawaybox}

\subsection{RQ3: What do agents lack when they miss the cause?}
\label{sec:genuine}
\label{sec:rq3}

Our evidence suggests that agents lack the simulator's output for each hypothesis, its signature (\cref{sec:example}). When the signatures are given, every model names the cause on almost every task.

The choice of probes is a small part of the failure. In \AuditHardProbeOpusTruthUniqueGivenEvidenceButWrong{} of Opus's \AuditHardProbeOpusFailures{} wrong answers in RQ2, the probes Opus had bought were already enough to single out the true hypothesis, as exact inference from the signatures shows (\cref{app:dt-audit}). We then removed probing. Every model is shown the same observations and must name the cause. Wherever the probes allow, only the true hypothesis has signatures equal to these observations (\cref{sec:benchmark}).

From the observations alone, the four models are right on \DPHaikuObs{} to \DPOpusObs{} of \DPN{} tasks (\cref{tab:rq2}). On pairs they are right on \DPHaikuObsPairs{} to \DPGptSixObsPairs{} of \DPPairsN. We then added the signature table, which lists what the simulator returns for every hypothesis on the same probes. For the running example the table would list each hypothesis's outcome on the initial mission. With the low heading gain alone the robot finishes that mission after 20\,s, and the true system needs 22\,s. The table turns the task into a lookup, which a plain string comparison solves on \SndStringCompareCorrect{} of \DPN{} tasks. Given the table, the four models are right on \DPHaikuSim{} to \DPOpusSim{} of \DPN{} tasks, and on \DPHaikuSimPairs{} to \DPOpusSimPairs{} of the \DPPairsN{} pairs.

The models do not supply the signatures themselves. Before seeing any observation, each model wrote a signature prediction for every hypothesis. Such a guess says, for example, how long the robot with only the low heading gain needs for the initial mission. These predictions rarely match the true signatures, and giving them to the model gives no detectable gain. Opus is right on \DPOpusOwn{} tasks with its own predictions and on \DPOpusObs{} without them (\cref{app:directpred}).

The two controls of \cref{sec:example}, run for Opus on every task, agree with this reading. The first control gives Opus every observation that the probes can return, with no budget. Its share of correct answers moves from \DiagHardProbeOpusAcc{} in RQ2 to \DiagHardFullobsOpusAcc. The second control is a tool that returns the simulator's output for any hypothesis on request. It makes Opus right on every task (\cref{app:st-rq3}). We read the diagnosis failure as lying in the step that needs the signatures, under a diagnosis prompt that does not say a cause may have two faults.

\begin{takeawaybox}\textbf{Takeaway (RQ3).} Shown the same observations, with no probes to choose, Opus, GPT 6.1, Sonnet and Haiku are right on \DPHaikuObs{} to \DPOpusObs{} of \DPN{} tasks. The simulator's output for each hypothesis raises them to \DPHaikuSim{} to \DPOpusSim, and their own predictions of that output give no detectable gain. We read the diagnosis failure as lying in the step that needs the simulator's output, under a prompt that does not say a cause may have two faults.\end{takeawaybox}

%% file: tables/main_repair.tex
\begin{tabular}{lrrrlrr}
\toprule
 & \multicolumn{4}{c}{verified repairs} & both faults & verified, \\
\cmidrule(lr){2-5}
model, standard tools & singles (44) & pairs (50) & rate (94) & 95\% CI & removed (pairs) & all 100 tasks \\
\midrule
Haiku 5.5 & 42/44 & 27/50 & 0.73 & [0.54, 0.90] & 9/27 & \HardAllStdHaikuFrac \\
Sonnet 5.5 & 44/44 & 48/50 & 0.98 & [0.93, 1.00] & 26/48 & \HardAllStdSonnetFrac \\
GPT-5.5 & 42/44 & 43/50 & 0.90 & [0.82, 0.98] & 23/43 & \HardAllStdGptFrac \\
GPT 6.1 & 42/44 & 49/50 & 0.97 & [0.92, 1.00] & 25/49 & \HardAllStdGptSixFrac \\
Opus 5.5 & 44/44 & 49/50 & 0.99 & [0.96, 1.00] & 17/49 & \HardAllStdOpusFrac \\
\bottomrule
\end{tabular}

%% file: tables/main_diag.tex
\begin{tabular}{lrrr}
\toprule
 & singles (44) & pairs (50) & all (94) \\
\midrule
\multicolumn{4}{l}{\emph{Reference procedures, given every hypothesis's signature}} \\
EIG oracle & 44/44 & 50/50 & 94/94 \\
random probing with exact inference & 22/44 & 47/50 & 69/94 \\
\midrule
\multicolumn{4}{l}{\emph{Agents that choose their own probes within the budget (RQ2)}} \\
Opus 5.5 & 32/44 & 10/50 & 42/94 \\
GPT-5.5 & 34/44 & 2/50 & 36/94 \\
Haiku 5.5 & 10/44 & 0/50 & 10/94 \\
\midrule
\multicolumn{4}{l}{\emph{No probing, the same observations for every model (RQ3)}} \\
Opus 5.5 & -- & \DPOpusObsPairs/\DPPairsN & \DPOpusObs/\DPN \\
GPT 6.1 & -- & \DPGptSixObsPairs/\DPPairsN & \DPGptSixObs/\DPN \\
Sonnet 5.5 & -- & \DPSonnetObsPairs/\DPPairsN & \DPSonnetObs/\DPN \\
Haiku 5.5 & -- & \DPHaikuObsPairs/\DPPairsN & \DPHaikuObs/\DPN \\
\midrule
\multicolumn{4}{l}{\emph{No probing, the same observations plus every hypothesis's signature (RQ3)}} \\
Opus 5.5 & -- & \DPOpusSimPairs/\DPPairsN & \DPOpusSim/\DPN \\
GPT 6.1 & -- & \DPGptSixSimPairs/\DPPairsN & \DPGptSixSim/\DPN \\
Sonnet 5.5 & -- & \DPSonnetSimPairs/\DPPairsN & \DPSonnetSim/\DPN \\
Haiku 5.5 & -- & \DPHaikuSimPairs/\DPPairsN & \DPHaikuSim/\DPN \\
\bottomrule
\end{tabular}

%% file: tables/fig_answers.tex
\begin{tikzpicture}[x=1mm,y=1mm,font=\scriptsize]
\definecolor{ansBoth}{HTML}{2A78D6}
\definecolor{ansBothExtra}{HTML}{EB6834}
\definecolor{ansOneAlone}{HTML}{1BAF7A}
\definecolor{ansOtherOne}{HTML}{EDA100}
\definecolor{ansNeither}{HTML}{E87BA4}
\node[anchor=east,font=\footnotesize] at (-1.5,-2.40) {Opus 5.5};
\fill[ansBoth] (0.20,0.00) rectangle (13.80,-4.80);
\node[inner sep=0pt,text height=1.45ex,text depth=0pt] at (7.00,-2.40) {\AnsOpusBoth};
\fill[ansBothExtra] (14.20,0.00) rectangle (23.60,-4.80);
\node[inner sep=0pt,text height=1.45ex,text depth=0pt] at (18.90,-2.40) {\AnsOpusBothExtra};
\fill[ansOneAlone] (24.00,0.00) rectangle (46.00,-4.80);
\node[inner sep=0pt,text height=1.45ex,text depth=0pt] at (35.00,-2.40) {\AnsOpusOneAlone};
\fill[ansOtherOne] (46.40,0.00) rectangle (50.20,-4.80);
\node[inner sep=0pt,text height=1.45ex,text depth=0pt] at (48.30,-2.40) {\AnsOpusOtherOne};
\fill[ansNeither] (50.60,0.00) rectangle (69.80,-4.80);
\node[inner sep=0pt,text height=1.45ex,text depth=0pt] at (60.20,-2.40) {\AnsOpusNeither};
\node[anchor=east,font=\footnotesize] at (-1.5,-9.60) {GPT-5.5};
\fill[ansBoth] (0.20,-7.20) rectangle (2.60,-12.00);
\node[inner sep=0pt,text height=1.45ex,text depth=0pt] at (1.40,-9.60) {\AnsGptFiveBoth};
\fill[ansOneAlone] (3.00,-7.20) rectangle (40.40,-12.00);
\node[inner sep=0pt,text height=1.45ex,text depth=0pt] at (21.70,-9.60) {\AnsGptFiveOneAlone};
\fill[ansOtherOne] (40.80,-7.20) rectangle (43.20,-12.00);
\node[inner sep=0pt,text height=1.45ex,text depth=0pt] at (42.00,-9.60) {\AnsGptFiveOtherOne};
\fill[ansNeither] (43.60,-7.20) rectangle (69.80,-12.00);
\node[inner sep=0pt,text height=1.45ex,text depth=0pt] at (56.70,-9.60) {\AnsGptFiveNeither};
\node[anchor=east,font=\footnotesize] at (-1.5,-16.80) {Haiku 5.5};
\fill[ansOneAlone] (0.20,-14.40) rectangle (26.40,-19.20);
\node[inner sep=0pt,text height=1.45ex,text depth=0pt] at (13.30,-16.80) {\AnsHaikuOneAlone};
\fill[ansOtherOne] (26.80,-14.40) rectangle (33.40,-19.20);
\node[inner sep=0pt,text height=1.45ex,text depth=0pt] at (30.10,-16.80) {\AnsHaikuOtherOne};
\fill[ansNeither] (33.80,-14.40) rectangle (69.80,-19.20);
\node[inner sep=0pt,text height=1.45ex,text depth=0pt] at (51.80,-16.80) {\AnsHaikuNeither};
\draw[black!45,line width=0.3pt] (0,-20.60) -- (70.00,-20.60);
\draw[black!45,line width=0.3pt] (0.00,-20.60) -- (0.00,-21.50);
\node[anchor=north,text=black!70,inner sep=1pt] at (0.00,-21.60) {0};
\draw[black!45,line width=0.3pt] (14.00,-20.60) -- (14.00,-21.50);
\node[anchor=north,text=black!70,inner sep=1pt] at (14.00,-21.60) {10};
\draw[black!45,line width=0.3pt] (28.00,-20.60) -- (28.00,-21.50);
\node[anchor=north,text=black!70,inner sep=1pt] at (28.00,-21.60) {20};
\draw[black!45,line width=0.3pt] (42.00,-20.60) -- (42.00,-21.50);
\node[anchor=north,text=black!70,inner sep=1pt] at (42.00,-21.60) {30};
\draw[black!45,line width=0.3pt] (56.00,-20.60) -- (56.00,-21.50);
\node[anchor=north,text=black!70,inner sep=1pt] at (56.00,-21.60) {40};
\draw[black!45,line width=0.3pt] (70.00,-20.60) -- (70.00,-21.50);
\node[anchor=north,text=black!70,inner sep=1pt] at (70.00,-21.60) {50};
\node[anchor=north,text=black!70,inner sep=1pt] at (35.00,-25.00) {number of two-fault tasks};
\fill[ansBoth] (75.00,-0.72) rectangle (78.00,-3.12);
\node[anchor=west,inner sep=0pt] at (79.20,-1.92) {both faults};
\fill[ansBothExtra] (75.00,-4.56) rectangle (78.00,-6.96);
\node[anchor=west,inner sep=0pt] at (79.20,-5.76) {both faults plus an extra hypothesis};
\fill[ansOneAlone] (75.00,-8.40) rectangle (78.00,-10.80);
\node[anchor=west,inner sep=0pt] at (79.20,-9.60) {one of the two faults alone};
\fill[ansOtherOne] (75.00,-12.24) rectangle (78.00,-14.64);
\node[anchor=west,inner sep=0pt] at (79.20,-13.44) {other answer with one of the faults};
\fill[ansNeither] (75.00,-16.08) rectangle (78.00,-18.48);
\node[anchor=west,inner sep=0pt] at (79.20,-17.28) {neither fault};
\end{tikzpicture}

%% file: sections/exp_real.tex
\subsection{RQ4: Do real bug-fix tasks separate the same agents?}
\label{sec:real}

Real bug-fix tasks separate the agents only in part. Haiku scores lowest, and the other three models are not ordered.

The tasks above are generated. As a check on real defects we ran the same repair agents on the real set. It has \RealN{} tasks built from merged bug-fix pull requests of ArduPilot, an autopilot, and of Nav2~\citep{macenski2020marathon} and tf2, two packages of the Robot Operating System (ROS). tf2 is the successor of the tf transform library~\citep{foote2013tf}. The agent gets the issue text and the repository just before the fix. A repair is graded by the test change of the same pull request, which the agent does not see (\cref{app:dt-real-setup}). The \emph{strict} reading counts a task only if the repair was verified in the run. The \emph{lenient} reading also counts a saved patch that failed in the run and was verified on one later re-grade at lower machine load.

\begin{table}[t]
\centering\small
\adjustbox{max width=\linewidth}{\input{tables/main_real}}
\caption{Repair on the \RealN{} tasks of the real set with standard tools, one attempt per task. The strict reading counts a task only if its repair was verified in the run. The lenient reading also counts a saved patch that was verified on one later re-grade. Haiku is lowest under both readings, and the other three models lie within a few tasks of each other.}
\label{tab:rq4}
\end{table}

Haiku, the weakest model on \hard{} (\cref{tab:rq1}), scores lowest under both readings (\cref{tab:rq4}). Under the strict reading Haiku's repair is verified on \RealStdHaikuVrr{} of the tasks, against \RealStdOpusVrr{} for Opus. Under the lenient reading the rates are \RealStdHaikuRetryVrr{} and \RealStdOpusRetryVrr. A paired test with Holm correction separates Haiku from each of the three stronger models under the lenient reading, and from Opus alone under the strict reading (\cref{app:dt-real-results}). The three stronger models lie within three tasks of each other under either reading, so the real set does not order them (\cref{app:dt-real-results}).

The real set differs from \hard{} in how repairs are graded, in which files the agent may edit and in the mix of tasks. The stronger models score lower on it than on \hard{}. Because of these differences, the lower scores do not show a loss of skill. Separately, the pull requests are public, so a model may have seen a fix in training, and our check for copied fixes cannot exclude that (\cref{app:dt-real-limits}).

\begin{takeawaybox}\textbf{Takeaway (RQ4).} Real bug-fix tasks separate the agents only in part. On \RealN{} real tasks Haiku scores lowest whether or not one re-grade is allowed, and the other three models are not ordered.\end{takeawaybox}

%% file: tables/main_real.tex
\begin{tabular}{lrrlrr}
\toprule
 & \multicolumn{3}{c}{strict reading} & \multicolumn{2}{c}{lenient reading} \\
\cmidrule(lr){2-4}\cmidrule(lr){5-6}
model, standard tools & verified & rate & 95\% CI & verified & rate \\
\midrule
Opus 5.5 & 88/97 & 0.91 & [0.83, 0.95] & 91/97 & 0.94 \\
GPT 6.1 & 85/97 & 0.88 & [0.80, 0.93] & 93/97 & 0.96 \\
Sonnet 5.5 & 85/97 & 0.88 & [0.80, 0.93] & 92/97 & 0.95 \\
Haiku 5.5 & 74/97 & 0.76 & [0.67, 0.84] & 81/97 & 0.84 \\
\bottomrule
\end{tabular}

%% file: sections/related.tex
\section{Related work}
\label{sec:related}

This section compares \bench{} with work on test-graded benchmarks, robot software, fault injection and diagnosis.

\para{Software-engineering benchmarks graded by tests.}
SWE-bench and its descendants grade an agent's patch by the repository's own tests~\citep{jimenez2024swebenchlanguagemodelsresolve,zheng2026swebenchproverifiedreliable}.
Audits find these verdicts often wrong~\citep{wang2025solvedissuesswebenchreally,yu2025utboostrigorousevaluationcoding,ahmed2026investigatingtestoverfittingswebench}.
Program-repair research had found the same problem earlier~\citep{smith2015cure,qi2015analysis}.
To our knowledge no version of SWE-bench covers a robotics repository.
A robot whose code passes every unit test can still drive off its route.
\bench{} grades a repair by how the simulated robot drives in hidden scenarios.

\para{Robot software and its testing.}
Bug studies catalogue the faults that occur in real robot software~\citep{timperley2024robust221bugsrobot,chen2025empiricalstudyinteractionbugs}.
Detectors find classes of these faults, such as mismatched physical units~\citep{ore2017phriky}.
Simulation and fuzzing check how robots behave~\citep{timperley2018crashing,kim2021pgfuzz}.
The closest prior work, RoboBugBench, grades LLM repairs of 33 real defects in a ROS~2 navigation stack in a physics simulator~\citep{robobugbench2026}.
RoboBugBench uses real defects. \bench{} injects its faults, so it controls each one.
It also adds a diagnosis track that asks for the cause without the code.

\para{Fault injection and mutation.}
Several works build tasks from synthetic bugs in real repositories that break an existing test~\citep{yang2025swesmithscalingdatasoftware,sonwane2025bugpilotcomplexbuggeneration}.
One of them, SWE-smith, already combines several bugs in one task.
Mutation testing injects small edits to assess a test suite~\citep{just2014mutants}.
Higher-order mutation combines several edits in one program~\citep{jia2009hom}.
\bench{} injects known faults, so RQ1 can check a verified repair for an injected edit left behind.

\para{Diagnosis and reference procedures.}
Root-cause benchmarks for agents mostly score how well an agent analyses telemetry that it is given~\citep{xu2025openrca,pham2024rcaeval}.
None of them compares the agent with a reference procedure.
Model-based diagnosis chooses each next measurement by expected information gain~\citep{dekleer1987gde,dekleer1989sherlock}.
To our knowledge no root-cause benchmark for agents combines priced probes, scoring against an exact-inference reference, two-fault causes, and controls that separate probe choice from signature prediction.
Our diagnosis track scores whether coding agents find the cause of code-level faults against such a reference (RQ2, RQ3).
\Cref{app:st-related} gives the full survey.

%% file: sections/discussion.tex
\section{Limitations}
\label{sec:discussion}

The limits below bear on the takeaways of \cref{sec:experiments}, and \cref{app:st-limits} lists every limit.

\para{Simulated robot, generated faults.}
RQ1 to RQ3 hold for generated faults in simulated navigation software.
The real set of RQ4 is a companion check. It does not validate the generator.

\para{The repair prompt, the README and the agent loop.}
The repair prompt asks for minimal fixes. A later run without that request removes both faults more often. Each gained full repair sets a gyro weight to a value other than the reference (\cref{sec:repair}).
Most faults left behind are constants whose correct values the README does not state, so leaving one in place need not mean a missed cause.
The agents run in our own agent loop. In a check with a shell agent on 20 of the 50 pairs, Opus's repair is verified on all 20 and removes both faults on 8. In our loop on the same pairs, 6 of 19 verified repairs remove both. The shell agent's two extra full repairs set a gyro weight to a value other than the reference (\cref{app:mv-miniswe}).

\para{Small numbers.}
With one attempt per task and few fault combinations, several intervals are wide. The models come from two vendors.
A difference of a few tasks between two runs can be noise.

\para{The diagnosis setup.}
The agent picks from a closed list, and the prompt never says that a cause may have two faults.
The low scores on two-fault tasks in RQ2 and in RQ3 without the signature table cannot be separated from a preference for single-fault answers.

\para{What the generator's files give away.}
A rule that only reads the generator configuration is right on \SndLookupAllRate{} of the main run, averaged over its random tie-breaks (\cref{app:dt-diag}).
No agent that chooses its own probes scores as high.
A program that undoes the edits it recognises from the operator library repairs \HardRecogVrr{} of the main run (\cref{app:dt-validation}).
A model that has seen the generator's configuration and operators should be tested on new fault combinations and held-out operators.

\section{Conclusion}
\label{sec:conclusion}

On generated robot navigation software, the repairs of Opus, GPT 6.1 and Sonnet pass every check on almost every task and often leave the quiet fault in place. Asked for the cause without the code, models rarely name both faults unless they are given the simulator's output for each hypothesis.
A benchmark for robot repair should report, beside the verified repair rate, whether a verified repair removes every injected fault and whether the agent can name the cause without the code. \bench{} reports both. For those who build diagnosis agents, our results suggest giving the agent a simulator it can query about its hypotheses.
On the \RealN{} tasks from real bug-fix pull requests, the weakest model on the generated tasks also scores lowest under both the strict and the lenient reading, and the other three models are not ordered.

%% file: sections/app_example.tex
\section{The running example: details}
\label{app:example}
\label{app:st-example}

This section gives the issue report, the reference repair and the catalogue of the running example of \cref{sec:example}, task \code{layered\_multi\_component\_11017}.

\para{Robotics terms.}
\begin{itemize}[leftmargin=*,itemsep=1pt,topsep=2pt]
\item A \emph{differential-drive} robot steers by the speeds of its two driven wheels.
\item \emph{Odometry} turns wheel-encoder ticks into an estimate of distance and heading change.
\item The \emph{pose} is the robot's position plus its heading.
\item The \emph{mission frame} is the coordinate frame of the waypoints. It may be shifted and rotated relative to the robot's start.
\item The \emph{watchdog} stops the motors when velocity commands stop arriving.
\end{itemize}

\para{Issue report.}
The README that the agent reads holds the issue report below. The generator wrote it from observed symptoms, and an audit found no word in it that names an injected fault (\cref{app:benchmark}).

\begin{examplebox}{Issue report (agent-visible, verbatim)}
\small
The robot leaves the planned route and ends far from the goal. Progress is made at first, then the trajectory deviates sharply. In some runs we instead see the following: waypoints are reached in the robot's own estimate, yet the physical positions visited do not match the mission. Changing the control or sensor update rates changes how strongly it manifests. Find the root cause in the software stack and fix it.
\end{examplebox}

\para{Reference repair.}
The reference repair is the generator's own correct fix. It has two one-line edits. The first restores the cap on the speed command in the motion controller. The second restores the heading gain in the configuration from 0.2 to 2.0.
\begin{lstlisting}
--- a/mbot/tracking.py
-        v = self.kp_distance * dist_goal
+        v = min(self.v_max, self.kp_distance * dist_goal)
--- a/mbot/config/defaults.json
-    "kp_heading": 0.2,
+    "kp_heading": 2.0,
\end{lstlisting}

\para{Catalogue.}
The diagnosis agent gets the twelve hypotheses below. Each software hypothesis carries the id of the operator that injects it. \code{K\_kp\_low} is the low heading gain, and \code{K\_v\_unclipped} is the missing speed cap. Entries marked [plant] are faults of the simulated robot itself, the plant of \cref{app:dt-plant}. The true entry is their composite, and the agent's list does not mark it.

\begin{examplebox}{Hypothesis catalogue (balanced, with the true entry unmarked)}
\small\raggedright
\code{K\_kp\_low} [software, configuration]: heading gain configured 10$\times$ too low.\quad
\code{K\_v\_unclipped} [software, motion controller]: linear speed command no longer limited to the configured maximum.\quad
\code{T\_watchdog\_unit} [software, actuator interface]: watchdog compares elapsed microseconds against a threshold scaled by 1000.\quad
\code{V\_arc\_displacement} [software, kinematics]: arc displacement re-implemented incorrectly.\quad
\code{K\_kp\_low+K\_v\_unclipped}, \code{K\_kp\_low+V\_arc\_displacement}, \code{T\_watchdog\_unit+K\_v\_unclipped}, \code{T\_watchdog\_unit+V\_arc\_displacement} [software, composite].\quad
\code{P\_sluggish\_motors} [plant]: motor controllers limit wheel acceleration to 6\,rad/s$^2$.\quad
\code{P\_wheel\_radius\_miscalibrated} [plant]: both wheels 10\% smaller than configured.\quad
\code{P\_encoder\_noise} [plant]: encoders report noisy tick counts.\quad
\code{no\_fault} [none]: the system has no defect.
\end{examplebox}

Each software operator id appears three times, once as a single and twice inside composites. Every composite has two components. For this reason the agent cannot find the true entry by counting how often a component appears or how many components a hypothesis has. The full menu of 27 probes can tell all twelve hypotheses apart.

%% file: sections/app_benchmark.tex
\section{Benchmark details}
\label{app:benchmark}

This section supports \cref{sec:benchmark} with the details of the robot, the tasks, the hidden checks and the scoring.

\subsection{The simulated robot and its software}
\label{app:dt-plant}
The \emph{plant} is the deterministic simulation of the robot, its sensors and its network. The robot has two driven wheels with speed and acceleration limits. Its sensors are wheel encoders, a noisy gyro for turn rate, a compass and an intermittent position fix. A simulated network link can delay, reorder and drop messages. The plant records the true pose and the applied wheel commands, and the software cannot write to this record.

Two correct software stacks can drive the plant. In the \emph{layered} stack, components exchange messages by topic through a rate-limited executor. This stack estimates the pose by fusing wheel motion with the gyro, and it corrects the heading with the compass. It steers toward a point ahead on the route. The \emph{callback} stack hands each message straight to its handler. It also fuses wheel motion with the gyro and steers along an arc to a point ahead, a method called pure pursuit.

\subsection{Operators and task generation}
\label{app:st-families}
\label{app:dt-operators}
An operator is a small edit to one module that still compiles and injects one fault. Each operator belongs to one mechanism family, the kind of mechanism it breaks. \cref{tab:families} lists the four families with example operators.

A \emph{masking operator} changes a configuration constant. It is harmless alone and lets another fault show, so the robot fails only when both faults are present. One masking operator weakens the compass correction, so the software trusts the compass too little. The robot still drives correctly with a correct gyro and fails only when the gyro is also faulty.

\begin{table}[t]
\centering\scriptsize
\input{tables/families}
\caption{Mechanism families, the kinds of mechanism an operator breaks, with two example operators each. Column ``ops'' counts each family's operators, \OpsTotal{} in all. SE(2) composition chains a shift and a rotation in the plane. Odometry is the wheel-measured motion. The watchdog stops the motors when commands stop. The executor delivers messages.}
\label{tab:families}
\end{table}

\para{Generation.}
For each candidate task the generator samples a software stack, a seed and one to three operators. It also picks one of several correct forms for each of \HealthyVariants{} functions that the operators do not touch. Every public identifier is then renamed. The \PublicScenarios{} public and \HiddenScenarios{} hidden scenarios are sampled independently of the fault. The issue report is written from observed symptoms. An audit rejects any report that names the operator, its target or an identifier from the edit.

\para{Acceptance.}
A candidate becomes a task only if every condition below holds. The correct renamed stack passes every hidden scenario. The faulty stack fails a behavioural requirement on a hidden scenario, or the metamorphic check, without raising an exception. The failure repeats on a second run. Renaming changes neither stack's scenario verdicts. The reference repair is non-empty. In a pair, some hidden scenario fails with both faults present.

\para{Selection of \hard.}
Apart from the eight replacement tasks (\cref{app:changes}), \hard{} was selected by structural criteria fixed before any model was run on it. Pairs had to be dominated by interacting scenarios, which fail only when both faults are present. Singles had to fail no public scenario or need a rating stated in the README for their fix. A rating is a rated limit such as the rated speed. A script applied our thresholds to the \SndSweepPairsScanned{} candidate pairs of sweeps that ran no model. The generator configuration lists the \SndRegenCompositionsN{} operator sets it kept.

The first generator run's acceptance conditions rejected every candidate of two of these sets. A separate generator run produced accepted pairs for one of the two. The six supplementary pairs come from that run and are reported apart from the main run.

\subsection{Hidden checks}
\label{app:dt-checks}
\label{app:oracles}
The verifier applies the agent's edits to a clean copy of the faulty code and swaps in its own simulator. The unit tests are the public tests plus hidden property tests for the geometry and kinematics modules. On each of the \HiddenScenarios{} hidden scenarios, the verifier evaluates the behavioural requirements of \cref{tab:oracles} from the plant's record. The scenarios vary the mission frame, sensor noise, network delay, control rate and robot geometry, and some cut the network link. The thresholds come from the verifier's own copy of the correct configuration. Edits to the software's copy of the limits leave them unchanged.

The metamorphic check~\citep{chen1998metamorphic} has \MetamorphicPairs{} pairs per task. Each pair takes a noise-free copy of one hidden scenario without a link cut and runs its mission twice, once as given and once with the whole world shifted and rotated. The second path, moved back, must match the first within 0.02\,m.
A repair is verified if every hidden scenario meets every requirement, every unit test passes, and the metamorphic check and the static configuration check hold.

\begin{table}[t]
\centering\scriptsize
\begin{tabular}{llp{0.56\linewidth}}
\toprule
group & requirement & rule (computed from plant truth and the verifier's healthy limits) \\
\midrule
protocol & no software error & the stack raised no exception \\
mission & mission completed & waypoints visited in order within 0.30\,m, last one before $\max(10, 1.5\,t_{\text{healthy}} + 4)$\,s; skipped on link-cut scenarios \\
mission & path tracking & max distance to the current segment $\le 0.50$\,m until completion or link cut \\
mission & final position / stopped at goal & final distance $\le 0.30$\,m and final speed $\le 0.05$\,m/s \\
estimation & position / heading consistency & max error of the software's logged estimates against interpolated truth $\le 0.3$\,m and $\le 0.3$\,rad (layered) or $0.4$\,rad (callback); fewer than six estimates fails \\
estimation & stamps monotonic / estimate rate & strictly increasing stamps; at least half of the expected odometry-rate count \\
safety & no plant violations & no non-finite or over-limit wheel command, robot stays in the workspace \\
configuration & ratings & configured maximum linear velocity (cruise speed) $\le 0.6$\,m/s and command timeout $\le 300$\,ms, read statically from the patched package \\
safety & linear / wheel speed limit & true speed $\le 1.10\times$ rated; commanded wheel speed $\le 1.01\times\min(\text{configured},\text{hardware})$ \\
safety & safety stop after link cut & stationary ($\|(v, 0.1\,\omega)\| \le 0.02$) by $t_{\text{cut}} + \text{timeout} + v_{\max}/(r\,a_{\max}) + 0.35$\,s, having moved before the cut \\
timing & command rate / actuation alive & logged commands per second within $[0.7, 1.6]\times$ control rate; plant-side wheel updates $\ge 0.7\times$ control rate \\
metamorphic & SE(2) equivariance & on noise-free copies of the scenario, the shifted and rotated mission's true trajectory, moved back, matches the base trajectory within 0.02\,m \\
\bottomrule
\end{tabular}
\caption{Hidden behavioural requirements, evaluated from the plant's record of true motion. Thresholds are fixed per task from the correct stack, whose completion time is $t_{\text{healthy}}$. The configuration row is the static check, read once from the patched package. The SE(2) row is the metamorphic check, evaluated on its \MetamorphicPairs{} pairs of a mission and a shifted and rotated copy. The other rows are evaluated on every hidden scenario.}
\label{tab:oracles}
\end{table}

\subsection{Repair and diagnosis scoring}
\label{app:dt-tracks}
\para{Repair.}
A verified pair repair counts as a full repair only if the injected text of both edits is gone from the patched module. An injected edit is \emph{README-determined} if it changes code logic whose correct form the README describes, or a constant whose value or bound the README states. A pair is README-determined if all its quiet faults are. The running example is README-determined, because its README states the rated speed that the missing cap breaks.

\para{Probes.}
A component's status report or the statistics of one message stream cost 1. A sensor's history or a mission under one of ten conditions, such as a long straight, cost 2. A test-bench drive on fixed commands costs 3. Observations are rounded to two significant figures and compared for exact equality.

\para{Correct answers.}
Let $C$ be the set of hypotheses that give the same observation as the true hypothesis under every probe. An answer is \emph{task-correct} if the set of hypotheses it names equals $C$. No answer gets partial credit.

\para{Budgeted probing and reference procedures.}
In budgeted probing the agent picks its own probes, with a summed price of at most \ProbeBudget, and then answers. The EIG oracle keeps the set $S$ of hypotheses whose signatures equal all observations so far, under a uniform prior. At each step it buys the probe $a$ that maximises
\[
\big(H(S) - \mathbb{E}_{o \sim a}[H(S \mid o)]\big)/\text{cost}(a),
\]
where $H$ is entropy. It stops when one hypothesis remains or no probe has positive gain. Random probing with exact inference buys probes uniformly at random, with one seeded draw per task, and updates $S$ in the same way.

\subsection{The direct-prediction study and the configuration-lookup rule}
\label{app:dt-rules}
\label{app:directpred}
The direct-prediction study removes probing. For each of the \DPN{} main-run tasks, $K$ is the set of probes that the EIG oracle bought. $K$ is empty when the initial mission alone singles out the true hypothesis. The target is the set of hypotheses that give the true observations on the initial mission and on $K$. Every prompt holds the README, the catalogue and the observations of the fault-free system on the same probes. No prompt shows code.

\begin{itemize}[leftmargin=*,itemsep=1pt,topsep=2pt]
\item \emph{Reference-probe observations.} The prompt adds the observations of the faulty system under diagnosis on these probes, and the model names the cause.
\item \emph{Signature table.} The prompt also adds the simulator's observation for every hypothesis on the same probes, as one static block of text.
\item \emph{Own signature predictions.} Before seeing any true observation, the model predicts every hypothesis's observation on these probes, \code{no\_fault} excepted. A later prompt adds these predictions to the reference-probe observations as the model's earlier estimates.
\end{itemize}

The plain string comparison calls no model. It keeps every hypothesis whose observations in the signature table, written as text, equal the true observations. It is scored by the task-correct rule.

Configuration lookup sends no probe. It reads only the agent-visible catalogue and the generator configuration. If the catalogue holds a composite whose operator set is one of the configuration's \SndRegenCompositionsN{} operator sets, it answers that composite. Otherwise it answers a software single whose operator the configuration lists as preferred, or any software single if none is listed. It breaks ties uniformly at random, and a pair catalogue can hold two configured composites. Its reported score is the expected task-correct score over these random choices.

%% file: tables/families.tex
\begin{tabular}{p{0.11\linewidth}rp{0.78\linewidth}}
\toprule
family & ops & example operators \\
\midrule
spatial & 45 & \code{S\_compose\_order}: SE(2) composition translates before rotating \newline \code{S\_odom\_halfbase}: Odometry heading increment divides by twice the wheel separation \\
temporal & 37 & \code{T\_watchdog\_unit}: Watchdog compares elapsed microseconds against a threshold scaled by 1000 \newline \code{T\_dt\_arrival}: Integration interval measured from arrival time instead of measurement stamps \\
communication & 26 & \code{B\_C\_twist\_key}: Controller emits angular velocity under a field the motors do not read \newline \code{C\_executor\_slow}: Executor spins at 20 Hz \\
control & 35 & \code{K\_v\_unclipped}: Linear speed command is no longer limited to the configured maximum \newline \code{BQ\_rated\_speed}: Cruise speed above the platform's rated 0.6 m/s (sampled) \\
\bottomrule
\end{tabular}

%% file: sections/app_validation.tex
\section{Validation of the verifier}
\label{app:validation}
\label{app:dt-validation}

This section supports the check of the verifier in \cref{sec:benchmark} and the last limit of \cref{sec:discussion}.

\para{Stored controls.}
Stored controls are repairs whose verdict is known in advance. They cover the \CtrlUniversalTasks{} generated tasks of \hard{} and of four earlier sets (\cref{app:mv-earlier}). Each task's reference repair, the generator's own correct fix, is checked twice. Five wrong repairs are applied to each task's faulty code. They are a patch that changes nothing, a patch that deletes tests, a loosening of the limits, an edit to the wrong subsystem and a random edit of a number. The verifier verified \CtrlReferenceOk{} of \CtrlReferenceN{} reference-repair checks. It rejected \CtrlNegativeOk{} of \CtrlNegativeN{} wrong repairs (\cref{tab:controls}). An equivalent repair is a different correct fix and should be verified. An overfit repair is correct only under public conditions and should be rejected. Repairs of both kinds exist for a few operators only. Some stored wrong-subsystem and random edits changed no file. These controls graded the unchanged faulty code.

Each of the five stored wrong repairs of the faulty code keeps the injected fault. The stored controls show that a repair which leaves the fault in place is rejected. They do not show that a wrong change to repaired code is rejected.

\begin{table}[t]
\centering\small
\adjustbox{max width=\linewidth}{\input{tables/controls}}
\caption{Stored controls per set. Each cell counts controls with the expected verdict over controls run. The reference repair is checked twice per task. The next five columns are wrong repairs of the faulty code. The text defines the equivalent and overfit repairs. Rows pilot (v1) and pilot v2 are the first and second pilots. Every cell should show zero mismatches.}
\label{tab:controls}
\end{table}

\para{Wrong edits on repaired code.}
On the main run we also combined the loosened-limits, wrong-subsystem and random edits with the reference repair (\cref{tab:ctl-repaired}). Each wrong edit was made to the faulty code as in the stored controls, and the reference repair was then applied on top of it. The loosened-limits edit was never verified. We sorted the \SndCtlHardAccTotal{} verified wrong-subsystem and random edits by inspection. Of these, \SndCtlHardAccClassNoop{} change no file and \SndCtlHardAccClassInert{} change only code the robot does not use, such as a status counter. Another \SndCtlHardAccClassWithinSpec{} change the robot's behaviour but meet every requirement the README states. The other \SndCtlHardAccClassFalseAccept{} are false accepts, wrong edits that break a README requirement and are verified.

In both false accepts on the main run, the edit makes the controller command a turn rate once the mission is finished. The robot then spins in place at the goal. The README asks the robot to stop there, but the check that it has stopped reads only its forward speed. The same gap gave \SndCtlCompAccClassFalseAccept{} false accept on compositions-100. A second gap appeared on the first pilot, where \SndCtlPilotAccClassFalseAccept{} edit disabled the branch of the controller that turns the robot in place. The hidden scenarios never reach that branch, because the generator samples start headings in a limited range. The README asks for correct behaviour from any start pose in the supported ranges and does not state those ranges.

\begin{table}[t]
\centering\small
\setlength{\tabcolsep}{3pt}\adjustbox{max width=\linewidth}{\begin{tabular}{@{}lccccccc@{}}
\toprule
 & \multicolumn{3}{c}{verified / graded} & \multicolumn{4}{c}{verified wrong-subsystem and random edits} \\
\cmidrule(lr){2-4}\cmidrule(lr){5-8}
set & loosened limits & wrong subsystem & random edit & no change & inert & within spec & false accept \\
\midrule
\hard{} main run & \SndCtlHardLoosenAcc/\SndCtlHardLoosenN & \SndCtlHardWrongAcc/\SndCtlHardWrongN & \SndCtlHardRandomAcc/\SndCtlHardRandomN & \SndCtlHardAccClassNoop & \SndCtlHardAccClassInert & \SndCtlHardAccClassWithinSpec & \SndCtlHardAccClassFalseAccept \\
\hard{} and four earlier sets & \SndCtlAllLoosenAcc/\SndCtlAllLoosenN & \SndCtlAllWrongAcc/\SndCtlAllWrongN & \SndCtlAllRandomAcc/\SndCtlAllRandomN & \SndCtlAccClassNoop & \SndCtlAccClassInert & \SndCtlAccClassWithinSpec & \SndCtlAccClassFalseAccept \\
\bottomrule
\end{tabular}}
\caption{Wrong edits combined with the reference repair, one grading each. The first three columns give verified over graded cases. Cases where the reference repair did not apply on top of the edit, because the edit overlaps the repair, are not graded. The last four columns sort the verified wrong-subsystem and random edits by inspection. The last column counts the false accepts.}
\label{tab:ctl-repaired}
\end{table}

\para{Operator-library recogniser.}
The recogniser holds the operator library. It searches a task's code for the replacement text of every operator, with identifiers as wildcards, and reverts each match. Its repair is verified on \HardRecogVrr{} of the main-run tasks. The recogniser and configuration lookup (\cref{app:dt-rules}) measure strategies that use the generator. They are not an upper bound on a model trained on the operator library.

\para{Other checks.}
One hidden test was replaced after it rejected a correct repair (\cref{app:changes}). A nearest-neighbour classifier that predicts a task's fault label from its sources was right on 0.27 of the first pilot, against 0.20 for always guessing the most common label. The pilot has five labels, the four families and a multi-component label, with six of its 30 tasks each. Grading the same submissions again reproduced every verdict.

%% file: tables/controls.tex
\begin{tabular}{lrrrrrrrrr}
\toprule
set & reference & noop & test deletion & loosen limits & wrong subsystem & random edit & equivalent & overfit & mismatches \\
\midrule
pilot (v1) & 60/60 & 30/30 & 30/30 & 30/30 & 30/30 & 30/30 & 5/5 & 1/1 & 0 \\
pilot v2 & 76/76 & 38/38 & 38/38 & 38/38 & 38/38 & 38/38 & -- & -- & 0 \\
compositions-100 & 200/200 & 100/100 & 100/100 & 100/100 & 100/100 & 100/100 & 2/2 & -- & 0 \\
interaction/spec & 56/56 & 28/28 & 28/28 & 28/28 & 28/28 & 28/28 & -- & -- & 0 \\
\hard{} main run & 188/188 & 94/94 & 94/94 & 94/94 & 94/94 & 94/94 & 14/14 & 12/12 & 0 \\
\hard{} supplementary pairs & 12/12 & 6/6 & 6/6 & 6/6 & 6/6 & 6/6 & -- & -- & 0 \\
\bottomrule
\end{tabular}

%% file: sections/app_repair.tex
\section{Results in detail: repair}
\label{app:repair-detail}
\label{app:extra}
\label{app:results}
This section supports the opening of \cref{sec:experiments} (Results) and RQ1 (\cref{sec:repair}).

\subsection{Run setup}
\label{app:dt-setup}
All models were called through one proxy under their aliases, so model snapshots are not pinned. No request set a temperature, a seed, or a reasoning or thinking parameter. The stored result rows record reasoning tokens only for GPT-5.5, non-zero in \SndReasonGptFiveRowsNonzero{} of its \SndReasonGptFiveRows{} rows. For the other models they record none, which can mean that none were used or that none were reported.

A repair agent works with one of two tool sets. With standard tools it can list, read, search and edit files, run the public tests, simulate the four public scenarios and submit. Robotics tools add the diagnosis probes, which measure the running robot, as tools. Repair grading stays the same.

Each model makes one attempt per task. An attempt is capped at \MaxSteps{} turns, \MaxWallS\,s and 2\,USD at list prices. GPT 6.1 has no entry in our price table, so its cap uses the prices of GPT-5.5. An attempt ends when the agent calls submit, reaches a cap or stops. The verifier grades the working copy as the agent left it, however the attempt ended.

Intervals are 95\% percentile bootstrap intervals over 2000 resamples of operator sets. An operator set is the operators a task is built from, one for a single and two for a pair. The intervals cover resampling of the observed operator sets. They do not cover new fault mechanisms or repeated attempts. Where every operator set has the same value the bootstrap collapses. \cref{tab:hard} then shows, with a dagger, the Wilson interval at the sample size corrected for clustering. Whether a pair is fully repaired depends almost entirely on its operator set. A paired test that treats generated tasks as independent, such as the exact McNemar test, overstates the evidence.

\subsection{Full repair table}
\label{app:st-fulltables}
\label{app:dt-repair}
\cref{tab:hard} gives every repair run on \hard{} with both tool sets, beside three rows without a model. The recogniser row is the operator-library recogniser of \cref{app:validation}. Adding the robotics tools changes the verified repair rate by $\HardRobMinusStdOpusDiff$ for Opus and by $\HardRobMinusStdHaikuDiff$ for Haiku. Neither change is significant under an exact McNemar test, with $p=\HardRobMinusStdOpusMcNemarP$ for Opus and $p=\HardRobMinusStdHaikuMcNemarP$ for Haiku. The supplementary column gives each run's rate on the six supplementary pairs. The no-op row shows that the unit tests alone accept the unrepaired repository on \HardNoopSwPassN{} of the 94 main-run tasks.

\begin{table}[t]
\centering\scriptsize
\adjustbox{max width=\linewidth}{\input{tables/hard100_repair}}
\caption{Repair on \hard{}. Singles and pairs count verified repairs. Main run is the verified repair rate (VRR) over 94. The 95\% CI is over operator sets. Behaviour-pass uses hidden scenarios only. Full repair counts pair repairs removing both edits. All tasks counts verified repairs over all 100 tasks. VRR is near 1.00 for most runs, and full-repair counts are low.}
\label{tab:hard}
\end{table}

\subsection{Which edits survive}
\label{app:mv-fullrepair-opset}
An injected edit still present after a verified repair is a surviving edit. Opus's verified pair repairs with standard tools leave \FrrHardStdOpusSurvTotal{} surviving edits. Of these, \FrrHardStdOpusSurvMasking{} come from masking operators (\cref{app:dt-operators}). One masking operator lowers the executor rate, how often the software's message loop runs. The other weakens how strongly the compass corrects the heading estimate. The README states neither constant, and the configuration that the agent sees holds only the injected value. The other \FrrHardStdOpusSurvTolerated{} surviving edits are edits that every evaluated check tolerates, such as the removed speed cap and a changed gyro weight.

\cref{tab:fullrepair-opset} splits the surviving edits of every run by operator. Whether an edit survives depends almost entirely on its operator. Over all ten runs, the executor-rate edit survives in \SndSurvAllRunsExecMid{} of the verified repairs of pairs that contain it. The compass-correction edit survives in \SndSurvAllRunsHeadCorr{} of such repairs. The speed-cap edit survives in \SndSurvAllRunsVUnclipped, although the README of every speed-cap pair states the rated speed that the cap enforces. The differences between models come from the two gyro-weight edits. GPT 6.1 and Sonnet often set a gyro weight to a value other than the reference, which counts as removed. If we count only verified pair repairs whose code at both edits equals the reference, Opus has 14 of 49, GPT 6.1 has 11 of 49 and Sonnet has 10 of 48.

Under the strict rule of \cref{app:dt-tracks}, \SndPairsReadmeDetStrict{} of the \SndPairsMain{} main-run pairs are README-determined. The lenient rule, shown in the table's lenient column, also counts the gyro weights and finds \SndPairsReadmeDetLenient{} such pairs. It counts them because the visible code fuses gyro and odometry headings with weights that sum to one, so a gyro weight above one gives odometry a negative weight. Which rule fits is a judgement, so the table shows both.

Some verified single-fault repairs also keep the injected text, \SndSinglesTextOpusStdKept{} of Opus's \SndSinglesTextOpusStdVerified{} verified singles with standard tools, mostly from operators that inject the same edit at several sites. A stricter rule counts a verified single as keeping the edit only if the function that the operator mutated is still identical to its faulty version after the repair. Under it Opus keeps the injected executor rate on \SndSinglesAstOpusStdKept{} verified singles with standard tools and \SndSinglesAstOpusRobKept{} with robotics tools.

\begin{table}[t]
\centering\scriptsize
\adjustbox{max width=\linewidth}{\input{tables/snd_fullrepair_opset}}
\caption{Full repair on the \SndPairsMain{} main-run pairs, by run. Columns give full over verified repairs, the clustered 95\% CI, surviving edits per operator, and full repairs on README-determined pairs under two rules. The last row counts surviving over verified instances in ten runs. The text discusses the first three operator columns.}
\label{tab:fullrepair-opset}
\end{table}

\subsection{The run without the request for minimal fixes}
\label{app:mv-ablation}
The repair prompt (\cref{app:prompts}) contains the sentence ``Prefer minimal, root-cause fixes over symptom suppression.'' We re-ran Opus with standard tools on the 50 main-run pairs with that sentence removed and everything else unchanged. The new run verifies \SndAblNewVerified{} of \SndAblNewN{} pairs and fully repairs \SndAblNewFull{} of them. The original run fully repairs \SndAblOldFull{} of its \SndAblOldVerified{} verified pairs. On the \SndAblBothVerified{} pairs verified in both runs, \SndAblGained{} gain a full repair and \SndAblLost{} lose one. The one pair that only the new run verifies is also fully repaired.

Each of the \SndAblGained{} gained full repairs removes a gyro-weight edit, and none of them restores the reference value. If we count only repairs that put the reference text back at every injected edit, the original run has 14 of 49 and the new run 14 of 50. Without the sentence the masking edits survive \SndAblNewSurvMasking{} times and the speed-cap edit \SndAblNewSurvVUnclipped{} times, as often as with it. An exact McNemar test on the gained and lost pairs gives $p=\SndAblMcNemarP$, unclustered. All gains come from operator sets that contain a gyro-weight edit, so this test overstates the evidence. Each task has one attempt in each run. The new run took place about a month after the original repair runs. A change in the served model between the two runs is not ruled out.

\subsection{The run with a shell agent}
\label{app:mv-miniswe}
Every other repair agent in the paper runs in our own agent loop, which offers a fixed set of tools as function calls and no shell (\cref{app:dt-setup}). To check whether this loop explains the surviving edits, we ran Opus under mini-swe-agent, version 2.2.7, a shell-only agent from the authors of SWE-agent~\citep{yang2024sweagent}. The agent works in a shell inside a container built from the sandbox image of the original runs, with no network. It can run the public tests and the simulator, and it can write its own scripts and scenarios. Its prompt keeps the text of the repair prompt, including the request for minimal fixes. An attempt may take up to 75 steps and 3\,USD. The tasks are 20 of the 50 main-run pairs, drawn at random with a fixed seed. Each task has one attempt, and the same verifier grades it.

Under the shell agent Opus's repair is verified on all 20 pairs and removes both injected edits on 8. In the original run on the same 20 pairs, 19 repairs are verified and 6 remove both. On each of the 12 pairs where the shell agent leaves an edit in place, the original run leaves the same edit. The shell agent's two additional full repairs both change a gyro-weight constant. In each the agent tries several values with a script of its own and keeps one that differs from the reference value. If we count only repairs that put the reference text back at every injected edit, the two runs agree, with 6 of 20 for the shell agent and 6 of 19 for the original run. The agent itself submitted 10 of the 12 attempts that leave an edit in place, and the other 2 ended at the cost limit. In all 12 the line that holds the edit appears in the agent's shell output, and no later command of the agent refers to it. The limits on an attempt do not explain these edits. Twenty pairs with one attempt each cannot show that the two loops give the same rate. The check gives no sign that our loop causes the surviving edits. As in the run without the request for minimal fixes (\cref{app:mv-ablation}), the original run is about a month older, and a change in the served model is not ruled out.

%% file: tables/hard100_repair.tex
\begin{tabular}{lrrrrlrrr}
\toprule
agent & singles (44) & pairs (50) & main run (94) & behaviour-pass & clustered 95\% CI & full repair (pairs) & supplementary (6) & all tasks (100) \\
\midrule
no-op (software tests pass) & 24/44 & 11/50 & 35/94 & -- & -- & -- & 0/6 & -- \\
operator-library recogniser & 35/44 & 44/50 & 0.84 & -- & -- & 44/44 & 1.00 & -- \\
reference repair & 44/44 & 50/50 & 1.00 & -- & -- & -- & 1.00 & -- \\
Haiku 5.5, standard tools & 42/44 & 27/50 & 0.73 & 0.90 & [0.54, 0.90] & 9/27 & 0.33 & \HardAllStdHaikuFrac \\
Haiku 5.5, robotics tools & 43/44 & 25/50 & 0.72 & 0.90 & [0.52, 0.90] & 9/25 & 0.50 & \HardAllRobHaikuFrac \\
Sonnet 5.5, standard tools & 44/44 & 48/50 & 0.98 & 0.98 & [0.93, 1.00] & 26/48 & 1.00 & \HardAllStdSonnetFrac \\
Sonnet 5.5, robotics tools & 44/44 & 49/50 & 0.99 & 0.99 & [0.97, 1.00] & 27/49 & 1.00 & \HardAllRobSonnetFrac \\
GPT-5.5, standard tools & 42/44 & 43/50 & 0.90 & 0.90 & [0.82, 0.98] & 23/43 & 1.00 & \HardAllStdGptFrac \\
GPT-5.5, robotics tools & 42/44 & 45/50 & 0.93 & 0.93 & [0.82, 0.99] & 26/45 & 1.00 & \HardAllRobGptFrac \\
GPT 6.1, standard tools & 42/44 & 49/50 & 0.97 & 0.98 & [0.92, 1.00] & 25/49 & 1.00 & \HardAllStdGptSixFrac \\
GPT 6.1, robotics tools & 43/44 & 49/50 & 0.98 & 0.99 & [0.94, 1.00] & 25/49 & 1.00 & \HardAllRobGptSixFrac \\
Opus 5.5, standard tools & 44/44 & 49/50 & 0.99 & 0.99 & [0.96, 1.00] & 17/49 & 0.83 & \HardAllStdOpusFrac \\
Opus 5.5, robotics tools & 44/44 & 50/50 & 1.00 & 1.00 & [0.96, 1.00]$^{\dagger}$ & 21/50 & 1.00 & \HardAllRobOpusFrac \\
\bottomrule
\end{tabular}

%% file: tables/snd_fullrepair_opset.tex
\begin{tabular}{lrlrrrrrrrr}
\toprule
 & & & \multicolumn{6}{c}{surviving edits by operator} & \multicolumn{2}{c}{README-determined pairs} \\
\cmidrule(lr){4-9}\cmidrule(lr){10-11}
run & full / verified & clustered 95\% CI & executor rate & compass corr. & speed cap & gyro (layered) & gyro (callback) & heading gain & strict & lenient \\
\midrule
Haiku 5.5, standard tools & \SndFrrHaikuStdFull/\SndFrrHaikuStdN & [\SndFrrHaikuStdClLo, \SndFrrHaikuStdClHi] & \SndFrrHaikuStdSurvExecMid & \SndFrrHaikuStdSurvHeadCorr & \SndFrrHaikuStdSurvVUnclipped & \SndFrrHaikuStdSurvGyroP & \SndFrrHaikuStdSurvGyroBP & \SndFrrHaikuStdSurvKpLow & \SndFrrHaikuStdDetStrictFull/\SndFrrHaikuStdDetStrictN & \SndFrrHaikuStdDetLenientFull/\SndFrrHaikuStdDetLenientN \\
Haiku 5.5, robotics tools & \SndFrrHaikuRobFull/\SndFrrHaikuRobN & [\SndFrrHaikuRobClLo, \SndFrrHaikuRobClHi] & \SndFrrHaikuRobSurvExecMid & \SndFrrHaikuRobSurvHeadCorr & \SndFrrHaikuRobSurvVUnclipped & \SndFrrHaikuRobSurvGyroP & \SndFrrHaikuRobSurvGyroBP & \SndFrrHaikuRobSurvKpLow & \SndFrrHaikuRobDetStrictFull/\SndFrrHaikuRobDetStrictN & \SndFrrHaikuRobDetLenientFull/\SndFrrHaikuRobDetLenientN \\
Sonnet 5.5, standard tools & \SndFrrSonnetStdFull/\SndFrrSonnetStdN & [\SndFrrSonnetStdClLo, \SndFrrSonnetStdClHi] & \SndFrrSonnetStdSurvExecMid & \SndFrrSonnetStdSurvHeadCorr & \SndFrrSonnetStdSurvVUnclipped & \SndFrrSonnetStdSurvGyroP & \SndFrrSonnetStdSurvGyroBP & \SndFrrSonnetStdSurvKpLow & \SndFrrSonnetStdDetStrictFull/\SndFrrSonnetStdDetStrictN & \SndFrrSonnetStdDetLenientFull/\SndFrrSonnetStdDetLenientN \\
Sonnet 5.5, robotics tools & \SndFrrSonnetRobFull/\SndFrrSonnetRobN & [\SndFrrSonnetRobClLo, \SndFrrSonnetRobClHi] & \SndFrrSonnetRobSurvExecMid & \SndFrrSonnetRobSurvHeadCorr & \SndFrrSonnetRobSurvVUnclipped & \SndFrrSonnetRobSurvGyroP & \SndFrrSonnetRobSurvGyroBP & \SndFrrSonnetRobSurvKpLow & \SndFrrSonnetRobDetStrictFull/\SndFrrSonnetRobDetStrictN & \SndFrrSonnetRobDetLenientFull/\SndFrrSonnetRobDetLenientN \\
GPT-5.5, standard tools & \SndFrrGptFiveStdFull/\SndFrrGptFiveStdN & [\SndFrrGptFiveStdClLo, \SndFrrGptFiveStdClHi] & \SndFrrGptFiveStdSurvExecMid & \SndFrrGptFiveStdSurvHeadCorr & \SndFrrGptFiveStdSurvVUnclipped & \SndFrrGptFiveStdSurvGyroP & \SndFrrGptFiveStdSurvGyroBP & \SndFrrGptFiveStdSurvKpLow & \SndFrrGptFiveStdDetStrictFull/\SndFrrGptFiveStdDetStrictN & \SndFrrGptFiveStdDetLenientFull/\SndFrrGptFiveStdDetLenientN \\
GPT-5.5, robotics tools & \SndFrrGptFiveRobFull/\SndFrrGptFiveRobN & [\SndFrrGptFiveRobClLo, \SndFrrGptFiveRobClHi] & \SndFrrGptFiveRobSurvExecMid & \SndFrrGptFiveRobSurvHeadCorr & \SndFrrGptFiveRobSurvVUnclipped & \SndFrrGptFiveRobSurvGyroP & \SndFrrGptFiveRobSurvGyroBP & \SndFrrGptFiveRobSurvKpLow & \SndFrrGptFiveRobDetStrictFull/\SndFrrGptFiveRobDetStrictN & \SndFrrGptFiveRobDetLenientFull/\SndFrrGptFiveRobDetLenientN \\
GPT 6.1, standard tools & \SndFrrGptSixStdFull/\SndFrrGptSixStdN & [\SndFrrGptSixStdClLo, \SndFrrGptSixStdClHi] & \SndFrrGptSixStdSurvExecMid & \SndFrrGptSixStdSurvHeadCorr & \SndFrrGptSixStdSurvVUnclipped & \SndFrrGptSixStdSurvGyroP & \SndFrrGptSixStdSurvGyroBP & \SndFrrGptSixStdSurvKpLow & \SndFrrGptSixStdDetStrictFull/\SndFrrGptSixStdDetStrictN & \SndFrrGptSixStdDetLenientFull/\SndFrrGptSixStdDetLenientN \\
GPT 6.1, robotics tools & \SndFrrGptSixRobFull/\SndFrrGptSixRobN & [\SndFrrGptSixRobClLo, \SndFrrGptSixRobClHi] & \SndFrrGptSixRobSurvExecMid & \SndFrrGptSixRobSurvHeadCorr & \SndFrrGptSixRobSurvVUnclipped & \SndFrrGptSixRobSurvGyroP & \SndFrrGptSixRobSurvGyroBP & \SndFrrGptSixRobSurvKpLow & \SndFrrGptSixRobDetStrictFull/\SndFrrGptSixRobDetStrictN & \SndFrrGptSixRobDetLenientFull/\SndFrrGptSixRobDetLenientN \\
Opus 5.5, standard tools & \SndFrrOpusStdFull/\SndFrrOpusStdN & [\SndFrrOpusStdClLo, \SndFrrOpusStdClHi] & \SndFrrOpusStdSurvExecMid & \SndFrrOpusStdSurvHeadCorr & \SndFrrOpusStdSurvVUnclipped & \SndFrrOpusStdSurvGyroP & \SndFrrOpusStdSurvGyroBP & \SndFrrOpusStdSurvKpLow & \SndFrrOpusStdDetStrictFull/\SndFrrOpusStdDetStrictN & \SndFrrOpusStdDetLenientFull/\SndFrrOpusStdDetLenientN \\
Opus 5.5, robotics tools & \SndFrrOpusRobFull/\SndFrrOpusRobN & [\SndFrrOpusRobClLo, \SndFrrOpusRobClHi] & \SndFrrOpusRobSurvExecMid & \SndFrrOpusRobSurvHeadCorr & \SndFrrOpusRobSurvVUnclipped & \SndFrrOpusRobSurvGyroP & \SndFrrOpusRobSurvGyroBP & \SndFrrOpusRobSurvKpLow & \SndFrrOpusRobDetStrictFull/\SndFrrOpusRobDetStrictN & \SndFrrOpusRobDetLenientFull/\SndFrrOpusRobDetLenientN \\
\midrule
all ten runs, surviving / verified instances & -- & -- & \SndSurvAllRunsExecMid & \SndSurvAllRunsHeadCorr & \SndSurvAllRunsVUnclipped & \SndSurvAllRunsGyroP & \SndSurvAllRunsGyroBP & \SndSurvAllRunsKpLow & -- & -- \\
\bottomrule
\end{tabular}

%% file: sections/app_diag.tex
\section{Results in detail: diagnosis}
\label{app:diag-detail}
This section supports RQ2 and RQ3 (\cref{sec:diag,sec:rq3}) and the last two limits of \cref{sec:discussion}. All results are on the 94 main-run tasks of \hard{} unless a sentence says otherwise.

\subsection{Full diagnosis table}
\label{app:dt-diag}
\cref{tab:harddiag} is the full version of \cref{tab:rq2}. It adds a row for configuration lookup and a column for the six supplementary pairs.

\begin{table}[t]
\centering\scriptsize
\adjustbox{max width=\linewidth}{\input{tables/hard100_diag}}
\caption{Diagnosis on \hard. Entries count task-correct answers (\cref{app:dt-tracks}) out of 44 singles, 50 pairs and six supplementary pairs, and a main-run rate over 94. Above the rule, policies choose probes, and EIG is expected information gain. Below it, every model gets the same observations. The pairs column carries the RQ2 comparison.}
\label{tab:harddiag}
\end{table}

Random probing with exact inference is right on fewer singles than Opus. In \HardSymptomAmbiguous{} of the \HardDiagBundles{} catalogues of \hard, supplementary pairs included, the initial observation does not single out the true hypothesis. On such a single, random probing often spends its budget before it picks the probe that tells the hypotheses apart.

In expectation, configuration lookup (rule in \cref{app:dt-rules}) is right on \SndLookupSinglesExp{} of \HardSingles{} singles and on \SndLookupPairsExp{} of \SndPairsMain{} pairs. Its expected rate over the main run is \SndLookupAllRate. On pairs and over the main run, this score is above random probing and above every agent that chooses its own probes.

The verifier compares observations with exact equality. If a numeric observation may differ from a signature by up to 10\%, the full menu of probes can still single out the true hypothesis in \TolHardIdentRelTen{} of the main-run catalogues, against \TolHardIdentexact{} under exact comparison. Opus's correct answers in budgeted probing then go from \TolHardOpusCorrectexact{} to \TolHardOpusCorrectRelTen, and its gap to the reference procedures stays in place. The tolerant count of \cref{tab:directpred}, \DPCeilTen, is a different quantity. It uses only the initial mission and the EIG oracle's probes, as in the direct-prediction study, and counts tasks on which the same 10\% rule applied to the simulator's signatures is task-correct.

Probing also differs in cost. On average the EIG oracle spends \SndOracleBudgetHard{} of the probe budget, and Opus spends \DiagHardProbeOpusBudget{} of it.

\subsection{Two controls for Opus}
\label{app:st-rq3}
The two controls of \cref{sec:example} were run for Opus on every main-run task (\cref{tab:controls-main}).
\begin{table}[t]
\centering\scriptsize
\adjustbox{max width=\linewidth}{\input{tables/diag_controls}}
\caption{Diagnosis controls for Opus on the 94 main-run tasks of \hard. Entries are task-correct rates (\cref{app:dt-tracks}). Each row says what the agent is given. The forward model reaches 1.00, and the full observation table is near budgeted probing.}
\label{tab:controls-main}
\end{table}
The full observation table gives Opus the true system's observation under every probe, with no budget, and asks for one answer. With it Opus is task-correct on \DiagHardFullobsOpusAcc{} of the main run, against \DiagHardProbeOpusAcc{} with budgeted probing. \DiagHardFullobsOpusFlipUp{} misses under budgeted probing become correct, and \DiagHardFullobsOpusFlipDown{} correct answers become wrong. The difference between the two rates is small. Whether it is significant depends on the test once tasks are clustered by operator set (\cref{app:dt-setup}).

The forward model keeps the budget and adds a tool, \code{predict(hypothesis, action)}, where the action is a probe. The tool returns the signature of any catalogue hypothesis under any probe. With it Opus is task-correct on \DiagHardForwardOpusAcc{} of the main run. \DiagHardForwardOpusFlipUp{} misses under budgeted probing become correct, and no correct answer becomes wrong. On the running example, the tool shows that with the low heading gain alone the robot completes the mission given in rotated coordinates. The true system, as Opus observed with its fifth probe, circles through 24\,rad and does not finish (\cref{tab:diagtrace}). With the forward model, Opus named both faults (\cref{sec:example}).

\subsection{The direct-prediction study}
\label{app:mv-controls-baselines}
\cref{tab:directpred} gives the results of the direct-prediction study of \cref{sec:rq3}. Its bottom row scores the signature table with no model, by exact comparison, by comparison within 10\% on numeric fields, and by comparison on categorical fields only. Its truth-kept column counts tasks where elimination with the model's own predictions within 10\% keeps the true hypothesis. In the worksheet arm the model gets the reference-probe observations and must mark every hypothesis consistent or inconsistent, citing the deciding field, before it answers.
\begin{table}[t]
\centering\scriptsize
\adjustbox{max width=\linewidth}{\input{tables/direct_prediction}}
\caption{Direct prediction on the \DPN{} main-run tasks of \hard{} (\DPPairsN{} pairs). Diagnosis columns count task-correct answers, pairs in parentheses. The worksheet arm asks the model to mark each hypothesis consistent or not. The last three columns score each model's own signature predictions against the simulator. The signature-table column gives the rates quoted in \cref{sec:rq3}.}
\label{tab:directpred}
\end{table}
Without the signature table, the four models name both faults on \DPHaikuObsPairs{} to \DPGptSixObsPairs{} of the \DPPairsN{} pairs. Adding the table raises this to \DPHaikuSimPairs{} to \DPOpusSimPairs. Only \DPHaikuCellTen{} to \DPGptSixCellTen{} of a model's predicted observations match the signature on every field within 10\%.

Given its own signature predictions beside the observations, Opus wins \DPOpusOwnVsObsWon{} tasks and loses \DPOpusOwnVsObsLost{} against the observations alone. An exact McNemar test finds no detectable gain for any of the four models ($p=\DPOpusOwnVsObsP$ for Opus, $p=\DPGptSixOwnVsObsP$ for GPT 6.1, $p=\DPSonnetOwnVsObsP$ for Sonnet, $p=\DPHaikuOwnVsObsP$ for Haiku).

The plain string comparison is task-correct on \SndStringCompareCorrect{} of \DPN{} tasks, so the signature table turns diagnosis into a lookup. Compared on categorical fields alone, such as whether a mission completes, the signatures single out the true hypothesis on \SndCatCeilAll{} of \DPN{} tasks. On pairs they do so on \SndCatCeilPairs{} of \DPPairsN.

\subsection{Failure audit}
\label{app:dt-audit}
We audited every failed submission of every reported run against the verifier's ground truth. For diagnosis, we replayed each agent's probes against the verifier's table of observations. The audit found no error in what the diagnosis agent is given. Every catalogue contains the true hypothesis, and every probe observation shown to Opus equals the verifier's entry.

\begin{table}[t]
\centering\scriptsize
\adjustbox{max width=\linewidth}{\input{tables/audit_diag}}
\caption{Wrong diagnoses classified by the agent's own evidence. Refuted answers contradict an observation the agent made, by a categorical field or by numbers only. Hedged answers add an excluded hypothesis to the true hypothesis. Consistent but incomplete answers agree with all the evidence and name only part of the hypotheses it cannot tell apart. The last column counts wrong answers made after the evidence singled out the true hypothesis.}
\label{tab:auditdiag}
\end{table}

\cref{tab:auditdiag} sorts the wrong answers into classes. Of Opus's \AuditHardProbeOpusFailures{} wrong answers, \AuditHardProbeOpusRefutedByOwnEvidence{} contradict an observation Opus itself made. In \AuditHardProbeOpusRefutedStrengthCategorical{} of these, the answer predicts a different mission outcome or state field, for example a completed mission where Opus saw it fail. \AuditHardProbeOpusHedgedSupersetIncludesTruth{} answers are hedges. \AuditHardProbeOpusEvidenceUnderdetermined{} answers are consistent but incomplete. They agree with all the evidence and name only part of the hypotheses it cannot tell apart.

In \AuditHardProbeOpusTruthUniqueGivenEvidenceButWrong{} of the \AuditHardProbeOpusFailures{} wrong answers, the probes Opus had bought were already enough to single out the true hypothesis. This class holds every hedge and all but one refuted answer. Exact inference on the probes Opus chose is task-correct on \DiagHardDecompOpusAcc{} of the main run. Opus loses accuracy after it has chosen its probes. For Haiku, most wrong answers also come after the evidence singles out the true hypothesis.

An unverified repair counts as leaving its fault in place if every hidden requirement it fails also fails without repair, or if the mutated function fails its hidden property test. Every unverified Opus repair on every set leaves its fault in place or is a regression, a failed requirement that the original defect did not break.
The audit found one defect in the hidden checks, a test that rejected a correct repair (\cref{app:changes}).

%% file: tables/hard100_diag.tex
\begin{tabular}{lrrrr}
\toprule
agent & singles (44) & pairs (50) & main run (94) & supplementary (6) \\
\midrule
EIG oracle & 44/44 & 50/50 & 1.00 & 1.00 \\
random probing, exact inference & 22/44 & 47/50 & 0.73 & 1.00 \\
configuration lookup, no probes (expected) & \SndLookupSinglesExp/\HardSingles & \SndLookupPairsExp/\SndPairsMain & \SndLookupAllRate & \SndLookupExtraPairsExp/\SndProvExtraAcc \\
Opus 5.5, budgeted probing & 32/44 & 10/50 & 0.45 & 0.67 \\
GPT-5.5, budgeted probing & 34/44 & 2/50 & 0.38 & -- \\
Haiku 5.5, budgeted probing & 10/44 & 0/50 & 0.11 & 0.17 \\
\midrule
\multicolumn{5}{l}{\emph{direct prediction, no probing}} \\
Opus 5.5, reference-probe observations & -- & \DPOpusObsPairs/\DPPairsN & \DPOpusObs/\DPN & -- \\
GPT 6.1, reference-probe observations & -- & \DPGptSixObsPairs/\DPPairsN & \DPGptSixObs/\DPN & -- \\
Sonnet 5.5, reference-probe observations & -- & \DPSonnetObsPairs/\DPPairsN & \DPSonnetObs/\DPN & -- \\
Haiku 5.5, reference-probe observations & -- & \DPHaikuObsPairs/\DPPairsN & \DPHaikuObs/\DPN & -- \\
Opus 5.5, signature table & -- & \DPOpusSimPairs/\DPPairsN & \DPOpusSim/\DPN & -- \\
GPT 6.1, signature table & -- & \DPGptSixSimPairs/\DPPairsN & \DPGptSixSim/\DPN & -- \\
Sonnet 5.5, signature table & -- & \DPSonnetSimPairs/\DPPairsN & \DPSonnetSim/\DPN & -- \\
Haiku 5.5, signature table & -- & \DPHaikuSimPairs/\DPPairsN & \DPHaikuSim/\DPN & -- \\
\bottomrule
\end{tabular}

%% file: tables/diag_controls.tex
\begin{tabular}{p{0.62\linewidth}r}
\toprule
Opus 5.5 run, and what the agent is given & \hard{} main run (94) \\
\midrule
budgeted probing: the probes it picks within the budget & \DiagHardProbeOpusAcc \\
full observation table: every probe's observation of the true system, no budget, one answer & \DiagHardFullobsOpusAcc \\
forward model: budgeted probing plus the signature of any catalogue hypothesis on request & \DiagHardForwardOpusAcc \\
\bottomrule
\end{tabular}

%% file: tables/direct_prediction.tex
\begin{tabular}{lrrrrrrr}
\toprule
 & \multicolumn{4}{c}{diagnosis correct (pairs)} & \multicolumn{3}{c}{own predictions} \\
\cmidrule(lr){2-5}\cmidrule(lr){6-8}
model & observations & + own predictions & + signature table & worksheet & cells within 10\,\% & categorical cells & truth kept (10\,\%) \\
\midrule
Claude Opus 5.5 & 38 (3) & 40 (7) & 94 (50) & 35 (4) & 0.15 & 0.71 & 0/94 \\
GPT 6.1 & 36 (5) & 39 (7) & 93 (50) & 30 (4) & 0.18 & 0.62 & 2/94 \\
Claude Sonnet 5.5 & 30 (2) & 28 (2) & 93 (50) & 30 (1) & 0.17 & 0.67 & 1/94 \\
Claude Haiku 5.5 & 27 (0) & 30 (1) & 89 (48) & 28 (0) & 0.14 & 0.66 & 1/94 \\
\midrule
plain string comparison & \multicolumn{4}{c}{94 (50)} & \multicolumn{3}{c}{tolerant elimination: 10\,\% 77, categorical only 19} \\
\bottomrule
\end{tabular}

%% file: tables/audit_diag.tex
\begin{tabular}{llrrrrrrr}
\toprule
set & run & $n$ & correct & \multicolumn{2}{c}{answer refuted by own evidence} & hedged & consistent but & truth unique \\
\cmidrule(lr){5-6} & & & & categorical & numeric only & (truth + extra) & incomplete & given evidence \\
\midrule
compositions-100 & Opus, probing & 100 & 48 & 9 & 16 & 11 & 16 & 36 \\
compositions-100 & Haiku, probing & 100 & 18 & 14 & 41 & 5 & 22 & 45 \\
compositions-100 & Opus, full observation table & 100 & 53 & 18 & 20 & 0 & 9 & 37 \\
interaction/spec & Opus, probing & 28 & 10 & 7 & 7 & 2 & 2 & 16 \\
interaction/spec & Haiku, probing & 28 & 6 & 6 & 12 & 1 & 3 & 18 \\
interaction/spec & Opus, full observation table & 28 & 12 & 9 & 6 & 0 & 1 & 14 \\
\hard{} & Opus, probing & 94 & 42 & 18 & 19 & 10 & 5 & 46 \\
\hard{} & Haiku, probing & 94 & 10 & 33 & 36 & 1 & 14 & 61 \\
\hard{} & GPT-5.5, probing & 94 & 36 & 25 & 26 & 1 & 6 & 52 \\
\hard{} & Opus, full observation table & 94 & 51 & 21 & 21 & 0 & 1 & 42 \\
\bottomrule
\end{tabular}

%% file: sections/app_real.tex
\section{The real set}
\label{app:real}
This section supports RQ4 (\cref{sec:real}) and the real set of \cref{sec:benchmark}.

\subsection{How the real set was built}
\label{app:dt-realset}
\label{app:dt-real-setup}
The real set has \RealN{} tasks from three open-source robot projects. Of these, \RealNArdu{} come from the ArduPilot autopilot, \RealNNav{} from Nav2~\citep{macenski2020marathon}, the navigation stack of the Robot Operating System (ROS), and \RealNTf{} from tf2, the ROS library that tracks coordinate frames and the successor of tf~\citep{foote2013tf}. Each task comes from a merged bug-fix pull request (PR). The agent gets an issue text and the repository at the commit just before the fix. It may read the whole repository and may edit only the directories that the fix touched. A backport is a task that applies an existing fix to another ROS~2 distribution, a separate release line. Of the tasks, 14 are backports (7 in Nav2 and 7 in tf2).

\para{Gates.}
A mined PR became a task only if its own test change fails at the commit before the fix and passes at the fix commit. The fix itself had to be verified and an empty patch rejected. For ArduPilot a fresh run had to reproduce the flip from failing to passing. A quality audit then removed \RealFunnelGateExcluded{} of the \RealFunnelGateAccepted{} tasks that passed these gates. Most were removed for an unfair hidden test or for a change that is not a behavioural repair.

\begin{table}[t]
\centering\small
\input{tables/real_funnel}
\caption{Mining of the real set by repository. The first column counts the curated candidate PRs. The other columns count tasks, and a backported fix gives several tasks per PR. ArduPilot has no backports, so its tasks are PRs. Its gate ran on the 56 curated candidates plus 72 PRs from a wider pool of merges that touch tests, 128 PRs in all. The tasks of the real set are those that passed the gate less those the quality audit removed. tf2 has no separate candidate list.}
\label{tab:realfunnel}
\end{table}

\para{Issue texts.}
On \RealCertRewritten{} of the \RealN{} tasks the issue text is a rewrite by Claude Opus 5.5 that describes only the symptom. On the rest it is the PR body or a linked issue. The rewriting model saw the PR, the linked issue and the failing test output. It also saw the names of the hidden tests and the diffs. It was told not to name the fix, the files it touches or the hidden tests. Model judges then checked each text for actionable symptoms and for leaks of the fix, with disagreements adjudicated by a further model pass. No human labelled the texts.

\para{The grader.}
The grader applies the agent's patch to a clean checkout of the commit before the fix. It takes the hidden test files from the fix commit, so an edit to a test cannot make it pass. It then builds the code and runs the PR's own test change. Every test that the fix turns from failing to passing must pass, and a fixed set of regression tests must not fail. ArduPilot tasks run a software-in-the-loop autotest, which drives the code against a vehicle simulator. Nav2 and tf2 tasks run unit tests built with the colcon tool. The agent loop and tools are those of the generated tasks (\cref{app:dt-setup}), with one attempt per task.

\subsection{Results}
\label{app:dt-real-results}
\cref{tab:real} gives each model's rates under both readings. The strict reading counts a task only if its repair was verified in the run. The lenient reading also counts a failed result whose saved patch was verified on one later re-grade at lower machine load. The \RealFailNoPatch{} failed results without a saved patch were not re-graded. Of \RealFailRegraded{} re-graded results, \RealFlipsRegraded{} were verified. Of the \RealFailRegression{} re-graded results that had failed the regression tests, \RealFlipsRegression{} were verified. We attribute these passes to load on the shared machine by exclusion, since the tests then passed in isolation. We later ran every test five times in isolation at the fix commit. None of the \RealFlakeHiddenInst{} hidden-test instances changed verdict, and \RealFlakeRegFlaky{} of \RealFlakeRegInst{} regression-test instances did.

\begin{table}[t]
\centering\scriptsize
\adjustbox{max width=\linewidth}{\input{tables/real_r2}}
\caption{Verified repair on the \RealN{} real-set tasks with standard tools. Column \hard{} repeats the main-run rate. Strict is the share verified in the run, with a 95\% interval. Lenient adds one re-grade of the saved patch. Non-backport is the strict rate on \RealCanonN{} tasks. Haiku is lowest under both readings.}
\label{tab:real}
\end{table}

The paired test is an exact McNemar test on the tasks where only one of two models is verified. Under the strict reading, after a Holm correction for three tests, it separates Haiku from Opus alone (adjusted $p=\SndRealMcNStrictOpusHolm$ for Opus and $p=\SndRealMcNStrictGptSixHolm$ for GPT 6.1 and Sonnet). Under the lenient reading it separates Haiku from each of the three stronger models, with largest adjusted $p=\RealMcNRetryHolmMax$. On the tasks that are not backports, the strict rates are \RealStdGptSixCanonVrr{} to \RealStdOpusCanonVrr{} for the stronger models and \RealStdHaikuCanonVrr{} for Haiku. The Wilson 95\% intervals treat tasks as independent, although each backport shares its fix with another task. Neither the strict nor the lenient rate is a bound on correctness.

\subsection{What limits the reading}
\label{app:dt-real-limits}
\para{Copied fixes.}
Because the PRs are public, a model may have seen a fix in training. We measured the share of a fix's distinctive added lines that a verified patch reproduces word for word. The mean share is \RealStdHaikuGoldRecall{} for Haiku and up to \RealStdOpusGoldRecall{} for Opus. The exact reproductions are one-line fixes with a single correct edit. For every model the share is lower on PRs merged before 2026 than on PRs merged in 2026 (\cref{tab:real-recall}). No model's training cutoff is recorded, so this split does not show that the 2026 PRs are unseen. It also cannot tell no recall apart from recall in both periods. The comparison has no test and is confounded with the size of the fix.

\begin{table}[t]
\centering\scriptsize
\adjustbox{max width=\linewidth}{\input{tables/snd_real_recall}}
\caption{Mean share of a PR fix's distinctive added lines that a verified patch reproduces, by the year the PR was merged. $n$ counts the verified patches scored. No test or interval.}
\label{tab:real-recall}
\end{table}

\para{The editable scope.}
We re-ran \RealScopeRepoN{} non-backport tasks that Opus had already verified with the whole repository editable. Opus verifies \RealScopeCtrlOpusVerified{} of them and GPT 6.1 \RealScopeCtrlGptSixVerified. Sonnet verifies \RealScopeCtrlSonnetVerified{} and Haiku \RealScopeCtrlHaikuVerified. Removing the limit changes little on these tasks. The tasks were drawn from Opus's successes, so the re-run does not measure the average effect of the limit. A second re-run on \RealScopeRandN{} non-backport tasks drawn at random found no difference between the two scopes that an exact McNemar test finds significant. Twenty tasks can rule out only a large effect.

\para{Run-to-run variance.}
The second re-run above also ran its \SndRealRepTasks{} random tasks for every model with the editable scope of the reported runs, limited to the fix's directories. These fresh attempts are the only repeats under the reported condition. They were graded by the final grader, which re-runs a failing test once in place, and with more builds running at once, so they are not a pure repeat. Their verdicts differ from the strict reading on \SndRealRepFlipStrict{} of \SndRealRepCells{} model-task results and from the lenient reading on \SndRealRepFlipLenient. The whole-repository runs of the two re-runs share some tasks, and on these \SndRealRepoRepFlip{} of \SndRealRepoRepCells{} results differ.

\para{Comparison with the generated tasks.}
The three stronger models score \RealDropOpus{} to \RealDropSonnet{} lower on the real set than on \hard{} under the strict reading. The two sets differ in the grader, the editable scope and the mix of tasks. The editable scope, the size of the fix and the hidden test may each affect the level of the scores. The lower rates cannot be read as a loss of skill.

%% file: tables/real_funnel.tex
\begin{tabular}{lrrrr}
\toprule
repository & candidate PRs & fail-to-pass gate run & gate passed & in the real set \\
\midrule
ArduPilot & 56 & 128 & 67 & 54 \\
Nav2 & 108 & 108 & 41 & 30 \\
tf2 & -- & 21 & 14 & 13 \\
\bottomrule
\end{tabular}

%% file: tables/real_r2.tex
\begin{tabular}{lrrrllrrrr}
\toprule
 & \hard{} & \multicolumn{5}{c}{real set} & \multicolumn{3}{c}{by repository} \\
\cmidrule(lr){3-7}\cmidrule(lr){8-10}
model & main run (94), std.\ tools & verified & strict & 95\% CI & lenient & non-backport & ArduPilot & Nav2 & tf2 \\
\midrule
Claude Opus 5.5 & 0.99 & 88/97 & 0.91 & [0.83, 0.95] & 91/97 (0.94; [0.87, 0.97]) & 0.90 & 48/54 & 27/30 & 13/13 \\
GPT 6.1 & 0.97 & 85/97 & 0.88 & [0.80, 0.93] & 93/97 (0.96; [0.90, 0.98]) & 0.87 & 50/54 & 22/30 & 13/13 \\
Claude Sonnet 5.5 & 0.98 & 85/97 & 0.88 & [0.80, 0.93] & 92/97 (0.95; [0.88, 0.98]) & 0.87 & 46/54 & 26/30 & 13/13 \\
Claude Haiku 5.5 & 0.73 & 74/97 & 0.76 & [0.67, 0.84] & 81/97 (0.84; [0.75, 0.90]) & 0.76 & 39/54 & 23/30 & 12/13 \\
\bottomrule
\end{tabular}

%% file: tables/snd_real_recall.tex
\begin{tabular}{lrrrr}
\toprule
model & before 2026, mean & before 2026, $n$ & 2026, mean & 2026, $n$ \\
\midrule
Opus 5.5 & \SndRealRecallPreOpus & \SndRealRecallPreNOpus & \SndRealRecallTwentySixOpus & \SndRealRecallTwentySixNOpus \\
GPT 6.1 & \SndRealRecallPreGptSix & \SndRealRecallPreNGptSix & \SndRealRecallTwentySixGptSix & \SndRealRecallTwentySixNGptSix \\
Sonnet 5.5 & \SndRealRecallPreSonnet & \SndRealRecallPreNSonnet & \SndRealRecallTwentySixSonnet & \SndRealRecallTwentySixNSonnet \\
Haiku 5.5 & \SndRealRecallPreHaiku & \SndRealRecallPreNHaiku & \SndRealRecallTwentySixHaiku & \SndRealRecallTwentySixNHaiku \\
\bottomrule
\end{tabular}

%% file: sections/app_history.tex
\section{Earlier sets and changes after the first results}

This section supports \cref{sec:example} and \cref{sec:experiments}. It describes the four earlier sets, one of which holds the running example, and lists the changes made after the first model results.

\subsection{Earlier generated sets}
\label{app:mv-earlier}
The same generator built four sets before \hard{}. On a first pilot of 30 tasks, 24 with one fault and 6 with two, the operator-library recogniser of \cref{app:validation} repaired 0.97 of the tasks. This result led us to weight the later sets toward tasks with several faults. A second pilot of \PilotVN{} tasks followed changes to the operators that made them harder to recognise. Compositions-100 has \CompSingles{} singles, \CompPairs{} pairs and \CompTriples{} tasks with three faults. Interaction/spec has \IspecSingles{} singles whose fix needs a rating stated in the README, and \IspecPairs{} pairs.

The running example is a pair of interaction/spec, and \hard{} holds pairs built from the same two operators. The running example is one of \SndIspecPairsSingleFails{} interaction/spec pairs in which one fault alone already fails a hidden scenario. In the other \SndIspecInteractionOnly{}, the interaction-only pairs, both faults are quiet, and neither fault alone fails a hidden scenario. With standard tools Opus's repair is verified on \MainIspecStdOpusVrr{} of interaction/spec (\cref{tab:repair}), including 11 of the set's \IspecPairs{} pairs. In diagnosis Opus names both faults on \DiagIspecProbeOpusTwoF{} pairs. The EIG oracle and random probing are each right on \DiagIspecRandomTwoF. Each of Opus's \SndIspecOpusPairsCorrect{} correct pair answers is on an interaction-only pair.

\begin{table}[t]
\centering\small
\adjustbox{max width=\linewidth}{\input{tables/repair_main}}
\caption{Repair on compositions-100 and interaction/spec, one attempt per task. VRR is the verified repair rate. SW is the share of repairs that pass every unit test. SW$\wedge\neg$V counts repairs that pass every unit test and fail the hidden checks. Repaired runtime marks runs made after the robotics-tool runtime was repaired (\cref{tab:changes}). The Haiku rows come from that separate run on the repaired runtime. They do not use Haiku's three repeated attempts of \cref{app:mv-variance}. The text discusses the Opus rows.}
\label{tab:repair}
\end{table}

\subsection{Changes after the first results}
\label{app:changes}
\cref{tab:changes} lists the changes made after a model had been run that affect a reported number. Changes to file names, metadata and documentation came after every reported run and change no input of one. A configuration check added to the verifier changed \CfgRegradeFail{} of the \CfgRegradeChecked{} stored verified repairs. A hidden test of pose integration was also rewritten, and it had rejected no model submission.

The selection rules of \hard{} were fixed before the first model run on it. Every later change to the protocols or the reference procedures is a row of \cref{tab:changes}. The masking operators (\cref{app:dt-operators}) were written after agents on the earlier sets had been seen to leave one of two faults in place. They supply most of the edits that survive Opus's verified pair repairs. The full-repair analysis, the diagnosis controls, the direct-prediction study, the tolerance replays of the reference procedures and the repair-prompt ablation were designed after the first results and are post hoc. The harder tier of real tasks (\cref{app:st-limits}) is post hoc as well. The choice of models was not fixed in advance for the diagnosis track (\cref{sec:diag}) or for the real set.

\begin{table}[t]
\centering\small
\begin{tabular}{@{}>{\raggedright\arraybackslash}p{0.36\linewidth}>{\raggedright\arraybackslash}p{0.33\linewidth}>{\raggedright\arraybackslash}p{0.25\linewidth}@{}}
\toprule
what changed & why & reported number it affects \\
\midrule
diagnosis catalogues re-balanced, every diagnosis condition re-run & their make-up gave the true composite away & every diagnosis number \\
eight \hard{} tasks replaced by tasks with logic faults & their fix value was printed in the README & every run on \hard \\
issue texts of \SndTextsChangedMain{} main-run tasks re-rendered, affected repair rows re-run & a model-judge audit found symptom terms that did not match the behaviour & repair of Opus, Haiku and GPT-5.5, Haiku's change with robotics tools \\
lookahead hidden test replaced by property checks, \RegradeN{} failed submissions re-graded, \RegradeFlipped{} of them verified & it rejected an equally correct ordering of the formula for the lookahead point, the route point the controller steers toward & repair verdicts on one task \\
robotics-tool runtime repaired, affected sets re-run & it failed \ToolErrPreFixPct\% of simulation requests & robotics-tools rates in \cref{tab:repair} \\
capped attempts graded from the working copy & they had counted as not submitted & 7 Opus repair rows \\
\midrule
check for a surviving edit made rename- and number-aware, configuration keys looked up through the identifier map & the old check matched canonical text only, and 5--8 edits per run were unresolved & every full-repair rate \\
\SndDiskFullFlippedToVerified{} Opus robotics-tools rows re-graded, later run again & their grading failed on a full disk & Opus's robotics-tools rate on \hard \\
ties of the reference procedures scored by the verifier's rule, no longer as $1/k$ & one rule for all rows & reference-procedure rates \\
list-valued fields of refuted answers typed element by element & one shared typing rule & refuted answers by type \\
\midrule
real-set texts rewritten as symptom reports, \SndRealExclN{} ArduPilot tasks excluded, Nav2, tf2 and three models added & re-validation, a quality audit and adjudication & every real-set number \\
lenient reading, each failed real-set row re-graded once at lower load & the regression tests proved load-sensitive & the lenient reading \\
two fail-to-pass tests removed & they pass at the bug commit in every run & real-set verdicts \\
four real-set texts rewritten and re-run, re-grade counts recomputed & a further model audit & real-set verified and re-grade counts \\
\bottomrule
\end{tabular}
\caption{Changes made after a model had been run that affect a reported number. Rows are grouped into task inputs and grading of \hard{}, scoring and accounting, and the real set. Superseded result files are kept.}
\label{tab:changes}
\end{table}

%% file: tables/repair_main.tex
\begin{tabular}{llrrrrrr}
\toprule
 & & \multicolumn{3}{c}{compositions-100 ($n$=100)} & \multicolumn{3}{c}{interaction/spec ($n$=28)} \\
\cmidrule(lr){3-5}\cmidrule(lr){6-8}
agent & tools & VRR & SW & SW$\wedge\neg$V & VRR & SW & SW$\wedge\neg$V \\
\midrule
no-op (unchanged repo) & -- & 0.00 & 0.08 & 8 & 0.00 & 0.57 & 16 \\
operator-library recogniser & -- & 0.85 & 0.88 & 3 & 0.79 & 0.79 & 0 \\
reference repair & -- & 1.00 & 1.00 & 0 & 1.00 & 1.00 & 0 \\
\midrule
Claude Opus 5.5 & standard tools (repaired runtime) & 0.91 & 1.00 & 9 & 0.96 & 0.96 & 0 \\
Claude Opus 5.5 & robotics tools (repaired runtime) & 0.97 & 1.00 & 3 & 0.96 & 1.00 & 1 \\
Claude Haiku 5.5 & standard tools (repaired runtime) & 0.86 & 0.92 & 6 & 0.68 & 0.79 & 3 \\
Claude Haiku 5.5 & robotics tools (repaired runtime) & 0.91 & 0.97 & 6 & 0.64 & 0.79 & 4 \\
\bottomrule
\end{tabular}

%% file: sections/app_related.tex
\section{Related work in full}
\label{app:st-related}
\label{app:compare}

This section supports \cref{sec:related} with the full survey of related work. \cref{tab:compare} sets \bench{} and its real set against the nearest benchmarks, with cells taken from the cited papers.

\begin{table}[t]
\centering\scriptsize
\setlength{\tabcolsep}{3pt}\begin{tabular}{@{}L{2.4cm}L{1.8cm}L{1.9cm}L{1.9cm}L{1.7cm}L{1.6cm}L{1.4cm}@{}}
\toprule
benchmark & defects & host code & behaviour-graded in simulation & hidden oracle & root cause scored & metered probes \\
\midrule
RoboBugBench~\citep{robobugbench2026} & 33 real & real ROS~2 & yes & yes & no & no \\
SWE-smith / PDB~\citep{yang2025swesmithscalingdatasoftware,chai2026precisedebuggingbenchmarkmodel} & generated & real repos & no & unit tests & edit precision & no \\
DDBench~\citep{yan2026evaluatingagenticcoderepair} & 60 real & real & no & reproducer & no & context arm \\
Active-SWE~\citep{li2026activeswebenchmarkingcodingagents} & real & real & no & unit tests & hunk match & unbudgeted \\
ROS Help Desk~\citep{katuwandeniya2025roshelpdeskgenai} & runtime-injected & real ROS & no & rubric & yes & unmetered \\
CaRE~\citep{hossen2023carefindingrootcauses} & configuration & real & partial & no & yes & no \\
RoboFuzz / PGFuzz~\citep{kim2022robofuzz,kim2021pgfuzz} & found by fuzzing & real & yes & policy oracles & no & no \\
FixDrive / DrPlanner~\citep{sun2025fixdrive,lin2024drplanner} & violations found in simulation & Apollo strategy DSL / planner heuristic and cost code, parameters & yes & scenario oracle & no & no \\
Multi-SWE-bench (C/C++)~\citep{zan2025multiswebench} & real & real repos & no & unit tests & no & no \\
Cloud-OpsBench / OpenRCA~\citep{wang2026cloudopsbench,xu2025openrca} & injected cloud faults & services & no & label match & yes & step cap, no oracle \\
\bench{} & generated, mechanism-typed & generated, renamed & yes (kinematic unicycle plant) & behavioural + metamorphic & greedy EIG reference, exact inference & budgeted, EIG reference \\
\bench{} real set & \RealN{} fix PRs (ArduPilot, Nav2, tf2) & real & SITL autotests for ArduPilot; gtest for Nav2/tf2 & fixing PR's tests & no & no \\
\bottomrule
\end{tabular}
\caption{How \bench{} and its real set differ from the nearest benchmarks, from the cited papers. No prior row combines generated code-level defects, hidden checks of simulated robot behaviour, exact root-cause scoring and metered probing. The real set is graded by the tests that each fixing pull request (PR) adds. SITL is software in the loop, the autopilot run against a vehicle simulator. gtest is the GoogleTest unit-test framework.}
\label{tab:compare}
\end{table}

\para{Software-engineering benchmarks graded by tests.}
SWE-bench and its descendants grade an agent's patch by the repository's own tests~\citep{jimenez2024swebenchlanguagemodelsresolve,zheng2026swebenchproverifiedreliable,pan2025trainingsoftwareengineeringagents,jain2025r2egymproceduralenvironmentshybrid,badertdinov2025swerebenchautomatedpipelinetask}.
A multilingual version adds tasks in C and C++~\citep{zan2025multiswebench}.
Audits find these verdicts often wrong~\citep{wang2025solvedissuesswebenchreally,yu2025utboostrigorousevaluationcoding,ahmed2026investigatingtestoverfittingswebench}. Program-repair research had found this earlier~\citep{smith2015cure,qi2015analysis}.

HWE-Bench grades repairs of hardware bugs with testbenches, the simulation harnesses of hardware design~\citep{cui2026hwebenchbenchmarkingllmagents}. DDBench compares agents given only the symptom with agents given extra context~\citep{yan2026evaluatingagenticcoderepair}. In Active-SWE the agent gets no issue report~\citep{li2026activeswebenchmarkingcodingagents}.
Interactive debugging is studied in its own environments and methods~\citep{yuan2025debuggymtextbasedenvironmentinteractive,kang2023explainableautomateddebugginglarge}. One work debugs multi-agent systems by intervening on their traces~\citep{ma2026doverinterventiondrivenautodebugging}.

To our knowledge no version of SWE-bench covers a robotics repository. None of these benchmarks asks the agent to name the cause. \bench{} grades a repair by how the simulated robot drives in hidden scenarios and adds a diagnosis track.

\para{Robot software and its testing.}
Bug studies catalogue the faults of real robot software. ROBUST lists 221 bugs in software for the Robot Operating System (ROS)~\citep{timperley2024robust221bugsrobot}. Other studies cover bugs in autonomous vehicles and drone autopilots~\citep{garcia2020avbugs,wang2021uavbugs} and bugs in how ROS components interact~\citep{chen2025empiricalstudyinteractionbugs}. Further studies cover fix patterns in two driving stacks, misconfigured ROS systems and drone flight logs with anomalies~\citep{chen2025adsbugfixpatterns,canelas2024rosmisconfig,wang2024uavloganomalies}.
We wrote the operators ourselves, without sampling them from the studied bugs, and then mapped them to the taxonomies of ROBUST, the interaction-bug study and the vehicle and drone studies.

Detectors find classes of these faults, such as mismatched physical units, wrong reference frames and misconfigured software architecture~\citep{ore2017phriky,kate2018phys,kate2021physframe,timperley2022rosdiscover}. One tool infers component models of ROS systems for such analysis~\citep{durschmid2024rosinfer}. One tool learns a model of normal behaviour from the telemetry of the ArduPilot autopilot and uses it to judge runs~\citep{afzal2022mithra}. Others repair misconfigured robot swarms or compare versions of one autonomous system~\citep{jung2021swarmbug,yousefizadeh2025cocomagic}. A study lists the challenges of testing robots~\citep{afzal2020challenges}. Robot repositories test their control code only superficially~\citep{chur2026beyondcontrolequations}.

Property-based testing, which checks a stated property on many random inputs, has been proposed for ROS~\citep{santos2018pbtros}. Control invariants derived from vehicle dynamics detect attacks at run time~\citep{choi2018controlinvariants}. Fuzzers feed robotic vehicles and ROS systems many generated inputs and flag unstable control, broken physical laws or violated safety policies~\citep{kim2019rvfuzzer,kim2022robofuzz,kim2021pgfuzz}. Other work generates stressful trajectories or lets agents test autopilots in simulation~\citep{hildebrandt2020stressful,duvvuru2025autosimtest}. Studies of ArduPilot and of navigation software find that most bugs can be reproduced in simulation~\citep{timperley2018crashing,sotiropoulos2017navigation}. Developers report that simulation is hard to use for testing~\citep{afzal2021simulation}. Metamorphic relations state how an output must change when the input changes in a known way~\citep{chen1998metamorphic}. They have been used to test autonomous systems~\citep{zhou2019metamorphicdriverless,lindvall2017metamorphic,tian2018deeptestautomatedtestingdeepneuralnetworkdriven,adigun2022metamorphictestingautonomoussimulations,ayerdi2024marmotmetamorphicruntimemonitoring,valle2026metamorphictestingvisionlanguageactionenabled}. In several works agents write new control code that is scored by running it, in simulation or against task metrics~\citep{ma2026rlebenchqualifyingexamcoding,elmaaroufi2026rhocodingagentsecretly,fu2026capxframeworkbenchmarkingimproving,shen2026embodiedswecodingagentslong,vandervleuten2025envtracesimulationbasedsemanticevaluation}.

The closest prior work, RoboBugBench, grades LLM repairs of 33 real defects in a ROS~2 navigation stack in a physics simulator. It reports repair rates of 24--36\%~\citep{robobugbench2026}. PGPatch repairs logic bugs in robotic vehicles from policy formulas, and PatchVerif measures how a patch changes the vehicle's physical state~\citep{kim2022pgpatch,kim2023patchverif}. FixDrive and DrPlanner repair driving behaviour and motion planners for automated vehicles~\citep{sun2025fixdrive,lin2024drplanner}.
Our hidden unit tests are property tests. Our metamorphic check requires the robot's path to shift and rotate with the mission when the whole mission is shifted and rotated in the plane. RoboBugBench uses real defects. \bench{} injects its faults, so it controls each one. None of these works checks whether the agent knows which fault caused the misbehaviour.

\para{Fault injection and mutation.}
Several works build debugging tasks from synthetic bugs in existing code~\citep{yang2025swesmithscalingdatasoftware,sonwane2025bugpilotcomplexbuggeneration,szalontai2026diffbasedcodecorruptionusing,chai2026precisedebuggingbenchmarkmodel}. SWE-smith and BugPilot keep bugs that break an existing test of a real repository. One of them, SWE-smith, already combines several bugs in one task. 

Synthetic bug sets are judged by how realistic they are and by whether training on them helps on real tasks~\citep{pham2025swesynth,chen2026sweatscale}. SWE-at-Scale reports the same solvers on synthesised and real tasks. It does not compare the two rankings.
Mutation testing injects small edits to assess a test suite~\citep{just2014mutants}. Higher-order mutation combines several edits in one program~\citep{jia2009hom}. Robot-specific mutation operators also exist~\citep{santos2025mutationtestingindustrialrobotic}. Mutation-based artificial bugs are also used to benchmark program repair~\citep{ouyang2024mubench}.
\bench{} injects known faults into a generated navigation stack with renamed modules and symbols. RQ1 can then check a verified repair for an injected edit left behind. The verifier also marks the hidden scenarios that fail only when both of a task's faults are present.

\para{Diagnosis and reference procedures.}
Model-based diagnosis compares what a system shows with what a model of it predicts~\citep{reiter1987theory,console1991spectrum,struss1989physical}. Its classical methods choose each next measurement by expected information gain~\citep{dekleer1987gde,dekleer1989sherlock}. Later work formalised this choice and extended it to unreliable tests~\citep{pattipati1990heuristic,raghavan1999unreliable}. More recent work makes sequential diagnosis scale~\citep{rodler2017generallyapplicablehighlyscalable,rodler2020dynamichsstreamliningreitershittingset}. Tables of the symptoms of each fault are older~\citep{bandler1985fault}. The methods have been applied to robot control software and ROS systems~\citep{steinbauer2005mbd,zaman2013diagnosis}, and fault detection and diagnosis for robots has been surveyed~\citep{khalastchi2018fault}.
Greedy choice of measurements is provably close to optimal under adaptive submodularity~\citep{golovin2017adaptivesubmodularitytheoryapplications}. We do not show that our objective meets those conditions and use the greedy policy only as a reference. Our requirement that the probes tell the true hypothesis apart is an analogue of diagnosability in discrete-event systems~\citep{sampath1995diagnosability}, and we do not reduce our setting to it.

Bayesian experimental design~\citep{rainforth2024modern} is now applied to how LLMs gather information~\citep{choudhury2026bedllmintelligentinformationgathering,hu2024uncertaintyofthoughts}. Agents that gather information have been scored by expected information gain or against an exact Bayesian reference, in games, supply chains and experimental design~\citep{grand2025shootfirst,deb2026stocktake,gandhi2025boxinggym}. These settings involve no code faults.

Most root-cause benchmarks for agents score the analysis of telemetry that the agent is given~\citep{xu2025openrca,fang2026openrca2,pham2024rcaeval,cai2026multidatasetbenchmarkevaluatingllm}. Others have the agent work on a running system with injected faults~\citep{jha2025itbench,clark2026sregym,chen2025aiopslabholisticframeworkevaluate}. One replays faults through diagnostic interfaces that the agent can query~\citep{wang2026cloudopsbench}. These benchmarks do not compare the agent with a reference procedure. Robot work searches driving systems for root causes, diagnoses faults injected at run time, ranks configuration options and scores root-cause text from issue tickets~\citep{zheng2026poirot,katuwandeniya2025roshelpdeskgenai,hossen2023carefindingrootcauses,herrmann2024diagnosingroboticssystemsissues}.
To our knowledge no benchmark combines priced probes scored against an exact-inference reference, scored two-fault causes, and controls that separate probe choice from signature prediction. Our diagnosis track scores coding agents on code-level faults against such a reference.

\para{Benchmark hygiene.}
Curation changes repair scores. About a third of the SWE-bench successes of SWE-agent with GPT-4 had a problem statement that leaked the solution~\citep{aleithan2024swebenchplus}. Rewriting problem statements as realistic user requests lowers scores and can reorder models~\citep{kim2026realswe}. Hints that point to the faulty code raise repair rates~\citep{alawad2026loc2repair,tamoyan2026sherloc}. ImpossibleBench measures how often agents exploit the tests~\citep{zhong2025impossiblebench}. Tests strengthened against adversarial patches lower the reported success~\citep{yu2026sweabs}. 

Agents retrieve upstream solutions or recall memorised code~\citep{ludwig2026shortcuttingfixidentifyingcategorizing}, and models show memorisation of benchmark repositories~\citep{liang2025swebenchillusionstateoftheartllms,dekoning2026metamorphictestingapproachdiagnosing,ramos2024memorizingbugbenchmarks}. Spaghetti Architect generates new code tasks with checked reference outputs~\citep{ji2026spaghettiarchitectcontaminationresistantbyconstructionlabelled}. SWE-rebench V2 certifies tasks with model judges~\citep{badertdinov2026swerebenchv2}. A review of benchmark practice gives recommendations for construct validity~\citep{bean2025constructvalidity}. How reliably benchmarks rank systems is itself studied~\citep{hardy2026agentevalreliability,patel2026predictivevalidity}.
We found no work that tests with a paired analysis whether models rank the same on generated and real tasks. Our real set also relies on model judges, and a further model pass adjudicates every disagreement (\cref{app:real}).

%% file: sections/app_limits.tex
\section{Limitations in full}
\label{app:st-limits}

This section lists every limit of the study, under the heads of \cref{sec:discussion} and three more.

\para{Simulated robot, generated faults.}
The simulated robot is a kinematic unicycle with no contact, perception, 3D, ROS or actuator lag. Faults that make a dynamic robot oscillate cannot be posed.
The two programs under test follow waypoints with one robot. The scores do not cover manipulation or perception code.
We wrote the operators and mapped them to four robot-bug taxonomies afterwards~\citep{timperley2024robust221bugsrobot,chen2025empiricalstudyinteractionbugs,garcia2020avbugs,wang2021uavbugs}. Several categories have no operator, and some message-bus operators have no ROS~2 counterpart.
The generator keeps only faults the simulator can expose. The operator mix does not show how often each bug occurs.
Templates write the issue reports.

\para{The verifier.}
A verified repair can keep an injected edit in a function that no hidden property test covers.
One hidden test rejected an equally correct repair, and the checks do not show that every correct alternative is verified.
A verified repair can leave the robot turning in place at the goal (\cref{sec:validation}) or fail from a start heading the hidden scenarios do not cover.

\para{The repair prompt, the README and the agent loop.}
The full-repair rate counts verified pair repairs in which the injected text of every edit is gone. An edit changed to a value other than the reference counts as removed. The rate is not a defect rate.
It is higher in a run a month later without the request for minimal fixes, where every gained full repair sets a gyro weight to a value other than the reference (\cref{app:mv-ablation}). A change in the served model is not ruled out.
Most edits left in place are configuration constants that the README neither states nor bounds, so they need not mean a missed cause.
The agents run in our own agent loop. A check with a shell agent on 20 of the 50 pairs gives no sign that this loop causes the edits left in place (\cref{app:mv-miniswe}).

\para{Small numbers.}
Models of two vendors, one attempt per task and few operator sets make the intervals, clustered by operator set, wide.
Verdicts change between repeated attempts, so a difference of a few tasks can be run-to-run variance.
Paired tests on the generated sets ignore the clustering, and their $p$-values overstate the evidence.
Each model is compared under its provider's default settings, and the stored records cannot show whether a model reasoned by default.

\para{The diagnosis setup.}
The catalogue is closed. Each operator appears in it equally often, yet its size still tells the agent how many faults to expect. The true cause is never a robot fault or \code{no\_fault}, the answer that nothing is wrong.
The reference procedures ignore the size cue and compare coarsened signatures by exact equality.
A lexical baseline that matches hypothesis descriptions to the initial observation is task-correct on \SndLexTaskCorrect/\LexHardN{} main-run tasks.
The prompt never says that a cause may have two faults. The low pair scores are not separated from a preference for single-fault answers.

The forward-model control bounds what an agent could do with a forward model. It does not measure unaided reasoning. Because it gives several aids at once, it cannot say which step Opus lacked.
The controls ran in full for Opus only. Haiku ran two of them on two earlier sets.
In the direct-prediction study, pair signatures identify the true hypothesis mostly through numeric fields that the README may not determine. Every arm sees the EIG oracle's probes.

\para{The real set.}
The real set covers three repositories, fixes whose pull request adds a test, four models and one agent scaffold.
The result holds for these \RealN{} tasks only.
Models rewrote and judged its issue texts with no human check. The editable scope names the directory of the fix.
The real set is a companion check. It does not validate the generator or stand alone as a benchmark of real debugging.
On the real set only the weakest model stands apart. Under the strict reading a paired test separates that model from Opus alone. A harder tier of \RealHardN{} real tasks with larger fixes, mined after the first results, does not show the same pattern, with \RealHardOpusFrac{} verified for Opus, \RealHardGptSixFrac{} for GPT 6.1, \RealHardSonnetFrac{} for Sonnet and \RealHardHaikuFrac{} for Haiku.
Each of the 14 backports shares its fix pull request with another task.
The fixes are public and no training cutoff is recorded, so the share of fix lines a patch reproduces does not bound contamination (\cref{sec:real}).

\para{Choices made after the first results.}
Several analyses and the masking operators were designed after the first results, so their results should be read as post hoc (\cref{app:changes}).
The choice of models was not fixed in advance. Two inexpensive models stopped after the first pilot, and GPT-5.5 did not run on the final real set.
The real set's tasks, issue texts and choice of models changed after earlier runs of Opus and GPT-5.5 on its ArduPilot tasks.
The eight replacement tasks of \hard{} meet neither structural selection criterion and lack the canary string that lets training corpora be checked.

\para{What the generator's files give away.}
Only a model whose training data predate 2026-10-13, the day the main run of \hard{} was generated, is certain not to have seen the set.
A new seed keeps the same operator combinations, so new combinations need new sweeps that run no model. Configuration lookup already beats random probing and every budgeted agent on pairs.
The existing sweeps hold no further pair outside the \SndRegenCompositionsN{} configured combinations that meets the selection rule.
We have not verified that regenerating \hard{} reproduces it bit for bit. The stored task files are the reference copy.

An adversary that compares a task's code with the template sources, and the nearest-neighbour classifier, were not run on \hard{}. Comparing a task's configuration with the template configuration is a shortcut for configuration faults that we have not tested. The recogniser and configuration lookup give no upper bound for a model trained on the generator.
The splits hold out operator combinations only. A model exposed to the tasks or the generator, or a claim about unseen mechanisms, needs new combinations and held-out operators.
Scores do not show that a deployed robot is safe. Unattended use would need a completion signal, a patch explanation and review before merging.
In the real-set task files we redacted mentions of people and an e-mail address after every reported run. Stored run records and mining caches still hold the original text.

%% file: sections/app_datasheet.tex
\section{Datasheet, prompts and reproducibility}
\label{app:datasheet}
\label{app:hosting}

This section supports \cref{sec:benchmark} and \cref{sec:real}. It gives the datasheet of \hard, the agents' prompts and tools, and how much repeated runs agree.

\subsection{Datasheet}

The datasheet follows the template of \citet{gebru2021datasheetsdatasets}.

\para{Motivation.} We built \hard{} to evaluate how coding agents repair and diagnose faults in robot navigation software. Whether such software is correct depends on how the robot behaves in closed loop, and unit tests alone do not decide it. There was no external funding specific to the dataset.

\para{Composition.} \hard{} has \HardN{} tasks. Of these, \HardLayered{} use the layered publish/subscribe architecture and \HardCallback{} the callback architecture. \HardSingles{} tasks are singles and \HardPairs{} are pairs, built from \HardOps{} distinct operators. Of the singles, \HardSpecSingles{} need a rating stated in the README and \HardPubInvSingles{} fail no public scenario. The other eight singles are the replacement tasks, which meet neither criterion (\cref{app:changes}). The singles fall into four mechanism families. Spatial, temporal and control faults account for \HardFamSpatial, \HardFamTemporal{} and \HardFamControl{} singles. Communication faults account for \HardFamComm.

Each task is a directory. Its \code{task.json} holds agent-visible metadata and the issue report. Its \code{diagnosis.json} holds the catalogue, the probe menu, the budget and the initial observation. The faulty repository with renamed modules and symbols, the simulator, four public scenarios, public tests and the README are in its \code{agent/} folder. A separate \code{verifier/} bundle holds the hidden checks, the reference patch and the ground truth, and it must never be mounted into an agent's environment. Every agent-visible repository holds a canary file, \code{CANARY.txt}, for contamination detection, except the eight tasks of the third generator run. The set holds no data about people and no offensive content.

\para{Collection process.} Three runs of the generator built the set, each with its own configuration and seed. The first run gave \SndProvMainAcc{} tasks, the second gave \SndProvExtraAcc{} supplementary pairs, and the third gave \SndProvReplAcc{} tasks that replaced \SndProvReplAcc{} tasks of the first run. The three runs drew \SndProvAllCand{} candidate tasks in all, and \cref{app:benchmark} gives the gates that accepted them. These candidate tasks differ from the \SndSweepPairsScanned{} candidate pairs of \cref{app:st-families}, which come from earlier sweeps that ran no model and from which a script chose the operator sets of the generator configuration. Each run's manifest records its configuration and seed. No human annotated any task.

\para{Preprocessing and labelling.} The task builder derives the labels from simulation and stores them in the verifier bundle. The labels are the fault operators, target modules, interaction classes, symptom classes, exposure conditions and hidden failure fractions. Issue reports are filled into templates from observed symptoms and were checked for leaked fault details.

\para{Uses.} \hard{} is meant for evaluating repair and diagnosis agents and for studying disagreement between checks. The generator can also produce training data, with the limit on comparability stated in \cref{sec:discussion}. The set must not be used to judge the safety of a deployed robot.

\para{Files and labels.} Each task has an agent-visible part, with \code{task.json}, \code{diagnosis.json} and \code{agent/}, and a verifier bundle that is kept apart for evaluators. Each task id names its fault family, and each \code{task.json} carries the fault families and difficulty metadata, so the agent-visible files reveal some labels. Issue texts are filled from templates and 97 of the \HardN{} are distinct. No word in an issue text names an injected fault, but within the set the issue text identifies the operator set for 96 tasks. The hash suffix of a task id also identifies the operator set. We keep the message-level trajectories of every reported repair run on \hard{} and the probe log and answer of every diagnosis attempt.

\para{Personal data.} The generated tasks hold no personal data. In the real-set task files, \SndRealHandleMentions{} mentions of people in \SndRealHandleFiles{} files and an e-mail address with a name in \SndRealEmailFiles{} file were redacted. The real set's run records and mining caches hold text from the upstream repositories, which may include personal data.

\subsection{Prompts and tools}
\label{app:prompts}

The system prompts of the repair agent, the diagnosis agent and the control that gives the full observation table are printed below as the agents received them.

\begin{lstlisting}[breakindent=0pt]
# Repair (both tool sets)
You are an expert robotics software engineer debugging a differential-drive robot's navigation stack.
You are given an issue report and a repository. Only files under the editable paths may be changed; `sim/` is the read-only hardware/physics simulation and `tests/`, `scenarios/` are read-only. Your fix must make the robot behave correctly for ANY robot geometry, start pose, mission frame, sensor/control rate and network condition within the supported ranges, not just for the public scenarios. Prefer minimal, root-cause fixes over symptom suppression. Do not loosen limits or disable checks to make tests pass. When you are confident, call submit_solution.

# Diagnosis (budgeted probing)
You are an expert robotics engineer diagnosing a robot that misbehaves. You cannot read the source code.
You can run diagnostic probes (each has a cost; you have a limited budget) and must then submit a structured diagnosis choosing the root cause from the hypothesis catalog. Observations are deterministic simulations rounded to two significant figures: the same fault always produces exactly the same observation, so any difference between an observation and what a hypothesis would produce, however small, is real. If several hypotheses remain indistinguishable with the available probes, list all of them as root_cause (a list) rather than guessing one.

# Diagnosis, full-observation-table control
You are an expert robotics engineer diagnosing a robot that misbehaves. You cannot read the source code.
You are given the COMPLETE set of diagnostic observations (every available probe has already been run) and a catalog of hypotheses. Infer the root cause. If several hypotheses are genuinely indistinguishable given the observations, list them all as root_cause; otherwise commit to one.
\end{lstlisting}

The two tool sets of the repair agent are described in \cref{app:dt-setup}. A diagnosis agent can only buy probes with \code{probe} and answer with \code{submit\_diagnosis}. The control that gives the full observation table offers only \code{submit\_diagnosis}. The forward-model control adds the following text to the user message.

\begin{quote}\small
You also have \code{predict(hypothesis, action)}: a simulator of the hypothesised fault that returns what the probe would show if that hypothesis were true (free). Compare predictions with real probe results to discriminate hypotheses.
\end{quote}

The diagnosis user message holds the README as the system description, the initial observation, the catalogue and the probe menu with its costs and budget. Every request carried the model alias, the messages, the tool schemas with \code{tool\_choice} set to \code{auto}, an output-token cap and a \code{user} field for attribution. The real-set prompt and the prompts of the direct-prediction study are not printed here.

\subsection{Run-to-run variance}
\label{app:mv-variance}

Some repeats come from earlier sets (\cref{app:mv-earlier}). Haiku's three repair attempts on compositions-100 give identical verdicts on \SndRepHaikuStdIdentical{} of \SndRepHaikuStdCommon{} tasks with standard tools. With robotics tools the count is \SndRepHaikuRobIdentical{} of \SndRepHaikuRobCommon. Opus's three diagnosis attempts on interaction/spec give the same task-correct verdict on \SndRepOpusIspecDiagIdentical{} of \SndRepOpusIspecDiagCommon{} tasks, and their answers agree on fewer tasks.

On \hard, attempts by Opus, GPT-5.5 and Haiku were re-run on the \SndTextsChangedMain{} main-run tasks whose issue texts were re-rendered. The input of these attempts changed, so their verdict changes bound sampling variance from above. Opus with standard tools went from verified to not verified on \SndRerunOpusStdTF{} of the \SndTextsChangedMain{} tasks and the other way on \SndRerunOpusStdFT. Haiku with robotics tools, the run that changed most, went from verified to not verified on \SndRerunHaikuRobTF{} tasks and the other way on \SndRerunHaikuRobFT. Sonnet and GPT 6.1 have no repeated attempts on \hard.